\documentclass[aoas]{imsart}

\usepackage{amsmath,amsfonts,bm}

\def\eqref#1{equation~\ref{#1}}

\def\1{\bm{1}}

\def\vtheta{{\bm{\theta}}}
\def\va{{\bm{a}}}
\def\vb{{\bm{b}}}

\def\vm{{\bm{m}}}

\def\vx{{\bm{x}}}

\def\mD{{\bm{D}}}

\def\mI{{\bm{I}}}

\def\mM{{\bm{M}}}

\def\mR{{\bm{R}}}

\DeclareMathAlphabet{\mathsfit}{\encodingdefault}{\sfdefault}{m}{sl}
\SetMathAlphabet{\mathsfit}{bold}{\encodingdefault}{\sfdefault}{bx}{n}

\newcommand{\E}{\mathbb{E}}

\newcommand{\R}{\mathbb{R}}

\newcommand{\Var}{\mathrm{Var}}

\RequirePackage[authoryear]{natbib}

\usepackage{microtype}
\usepackage{graphicx}
\usepackage{subcaption}
\usepackage{booktabs} 
\usepackage{multirow}

\usepackage{mathtools}

\usepackage[textsize=tiny]{todonotes}
\usepackage[utf8]{inputenc} 
\usepackage[T1]{fontenc}    
\usepackage{url}            
\usepackage{booktabs}       
\usepackage{amsfonts}       
\usepackage{nicefrac}       
\usepackage{microtype}      
\usepackage{xcolor,graphicx}         
\usepackage{amsthm,amssymb,amsmath,bm}
\usepackage{algorithm}

\usepackage{algorithmic}
\usepackage[algo2e]{algorithm2e}

\usepackage{wrapfig}
\usepackage{comment}
\usepackage{url}
\usepackage{caption}
\usepackage{placeins}

\newcommand{\bmu}{{\bm \mu}}
\newcommand{\bgamma}{{\bm \gamma}}
\newcommand{\bSigma}{{\bm \Sigma}}
\newcommand{\bsigma}{{\bm \sigma}}
\newcommand{\bPsi}{{\bm \Psi}}

\newcommand{\bF}{{\bm F}}

\newcommand{\bI}{{\bm I}}

\newcommand{\cI}{{\mathcal I}}
\newcommand{\cN}{{\mathcal N}}
\newcommand{\cP}{{\mathcal P}}

\newcommand{\bbE}{{\mathbb E}}
\newcommand{\bbI}{{\mathbb I}}
\newcommand{\bbP}{{\mathbb P}}

\def\red{\color{red}}

\SetKwInput{KwInput}{Input}   
\SetKwInput{KwOutput}{Output} 
\graphicspath{{figs/}}

\theoremstyle{plain}
\newtheorem{theorem}{Theorem}[section]

\newtheorem{lemma}[theorem]{Lemma}

\theoremstyle{definition}
\newtheorem{definition}[theorem]{Definition}

\theoremstyle{remark}

\begin{document}

\begin{frontmatter}
\title{MANDERA: Malicious Node Detection in Federated Learning via Ranking}
\runtitle{MANDERA: Malicious Node Detection in Federated Learning via Ranking}

\begin{aug}
\author[A]{\fnms{Wanchuang}~\snm{Zhu}\ead[label=e1]{wanchuang.zhu@sydney.edu.au}},
\author[B]{\fnms{Benjamin Zi Hao}~\snm{Zhao}\ead[label=e2]{ben\_zi.zhao@mq.edu.au}},
\author[C]{\fnms{Simon}~\snm{Luo}\ead[label=e3]{simon.luo@unsw.edu.au}},
\and 
\author[E]{\fnms{Ke}~\snm{Deng}\ead[label=e5]{kdeng@tsinghua.edu.cn}}\footnote{Corresponding author.}

\address[A]{University of Sydney, Australia \printead[presep={,\ }]{e1}}

\address[B]{School of Computing, Macquarie University, Australia \printead[presep={,\ }]{e2}}

\address[C]{School of Computer Science and Engineering, The University of New South Wales, Australia \printead[presep={,\ }]{e3}}


\address[E]{Department of Statistics and Data Science, Tsinghua University, China \printead[presep={,\ }]{e5}}
\end{aug}

\maketitle

\begin{abstract}
 
While federated learning is a popular framework for distributed learning in the machine learning community that allows a global model to be trained across decentralized devices without data exchanging, it is vulnerable to Byzantine attacks where some involved devices are manipulated to poison the model training. Defending federated learning from various Byzantine attacks has been an active research topic in machine learning in recent years.
This paper proposes a novel defense strategy called MANDERA which achieves effective defense via precise detection of the manipulated devices based on a statistical analysis of a ranking matrix obtained from the messages reported by decentralized devices.
Compared to existing defense strategies, MANDERA enjoys a higher defense efficiency against a wide range of Byzantine attacks and a clear theoretical guarantee.
The effectiveness and robustness of MANDERA are further confirmed by a collection of real data analyses.

\end{abstract}

\end{frontmatter}

\section{Introduction} \label{sec:intro}
Due to the extremely large data size and concerns on data privacy and security in modern data science problems, decentralized distributed learning has become a popular choice in statistical analysis and machine learning in recent years.
Under such a setting, we assume that data from different sources, which are referred to as \emph{local nodes} in the rest of the paper, cannot be pooled together for data analysis, and we have to train statistical or machine learning models based on separated datasets distributed in the participating local nodes with limited information transmission.


In statistical literature, distributed learning has been widely used for statistical inference of various statistical models under the above scenarios, including sparse regression \citep{lee2017communication}, isotonic regression \citep{baberjeeAOS2019}, quantile regression \citep{volgushev2019distributed}, ridge regression \citep{dobriban2020wonder}, general linear regression \citep{24GLMDE}, sparse \textit{M}-estimation \citep{tu2023byzantine}, and statistical optimization \citep{chen2022first,wu2023QuasiNewton,tang2024multi}.
The main idea is a \emph{divide-and-conquer} (DC) strategy obtains an individual estimator of the model parameters in each local node first, and then constructs an assembled estimator by aggregating the local estimators \citep{chen2014split,zhang2015divide,kleiner2014scalable,zhao2016partially,xichenAOS24}.
Typically, these methods can guarantee data privacy and security, because only summary statistics of local datasets are transmitted.

\begin{figure}
    \centering
    \includegraphics[width=0.5 \linewidth]{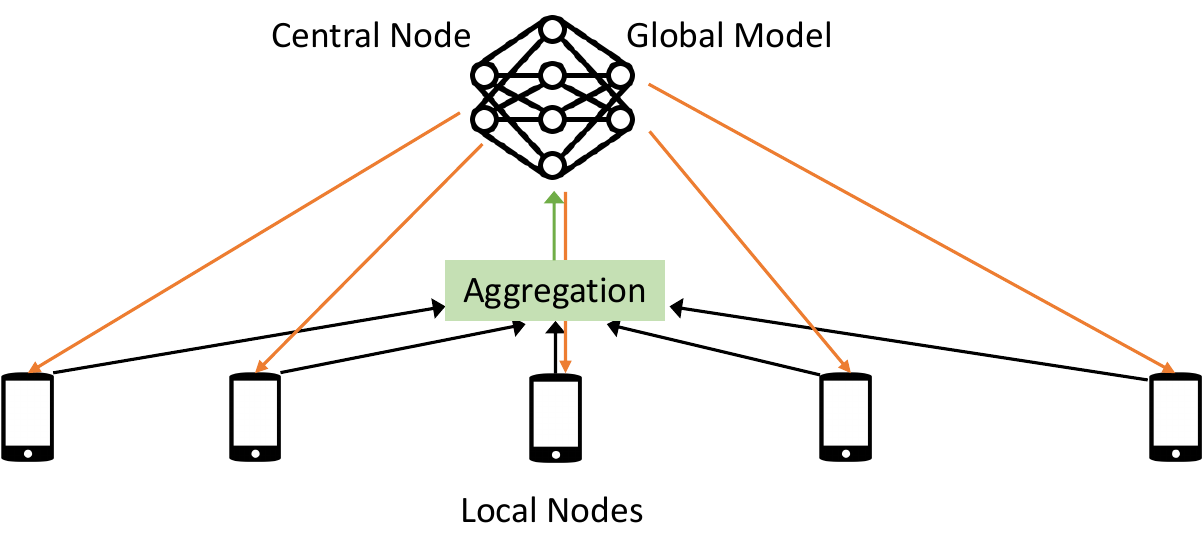}
    \caption{The general framework of federated learning with a central node equipped with a global model and a collection of local nodes holding distributed datasets. 
    The orange lines demonstrate the actions to send the global model to the local nodes. 
    The black lines represent the actions to send back the proposed updates of the global model obtained from local training in local nodes to the central node. The green line represents the action of aggregating the proposed updates from local nodes to a global update, and then sending it to the central node for global model updating.}
    \label{fig:WF}
\end{figure}

In machine learning literature, a closely related learning framework known as \emph{federated learning} (FL) has been established to train more complicated models, often deep neural networks, when data are distributed in local nodes \citep{mcmahan2017communication}.
Figure \ref{fig:WF} visualizes the general framework of FL, with the detailed workflow summarized below:
\begin{enumerate}
    \item Initialization: a global model is initialized on the central node;
    \item Local training: the global model is sent to individual local nodes, where local training takes place to propose an update to the global model;
    \item Model updates: instead of sending raw data to the central node, the local nodes send only the model updates (i.e., the proposed change to the model parameters) back to the central node;
    \item Aggregation: the central node aggregates the updates from all participating local nodes to update the global model, usually with techniques like averaging or weighted averaging; 
    \item Iteration: steps 2-4 are repeated for multiple iterations until the global model converges or reaches a satisfactory level of performance. 
\end{enumerate}
In real life, FL has been widely used to achieve distributed learning in many important problems and applications. For example, \cite{mcmahan2017communication} pointed out that image classification and language models are two ideal scenarios where federated learning can be applied.
In industry, high-tech companies, such as Google and Apple, have used FL widely in their products \citep{kairouz2021advances}. 
In medical image analysis, FL is very helpful for bypassing the dilemma of integrating medical images of patients across different medical centers due to privacy concerns \citep{guan2024federated}.
In these applications involving complex text or image data, the global model in the central node typically involves a deep neural network with tens of thousands of parameters, according to \cite{tolpegin2020data}.
In this case, model parameters are typically optimized via gradient-based method, e.g., stochastic gradient decent, and the key messages transmitted from local nodes to the global node become the gradient of parameters instead of the data.

Apparently, FL enjoys the common advantages shared by all distributed learning methods: an improvement in privacy protection because we do not need to pool together data distributed in those local nodes, and a decomposition of computation burden because the involved computations are offloaded to contributing local nodes rather than being concentrated solely on the central node. 
Compared to distributed learning methods in statistical literature for parsimonious statistical models, however, FL in machine learning literature is unique in two aspects:  
first, FL concerns models of much larger scale, often deep neural networks with extremely high dimensional parameters; 
second, FL often faces more complex operating environment due to the possible presence of malicious attackers who try to disturb the learning system with misleading messages.
However, with the presence of malicious attackers, FL is vulnerable to Byzantine attacks \citep{Krum2017,lamport2019byzantine,tolpegin2020data}, where some local nodes are manipulated by attackers to poison the global model in the central node with misleading messages~\citep{chen2017,baruch2019little, fang2020local, tolpegin2020data}, or backdoor attacks ~\citep{bagdasaryan2020backdoor} where some specific malicious functionality is inserted into the global model through poisoned messages from malicious local nodes and waken by a pre-defined trigger.

Here, we focus on the Byzantine attacks, and refer to the nodes manipulated by attackers as malicious nodes, and the other normal nodes as benign nodes. When the global model in the central node is a deep-learning-based model relying on gradient-based methods for model training, a typical malicious node in a Byzantine attack would send a poisoned fake gradient value to the central node to disturb the parameter updating of the global model.
If the poisoned gradient values from these manipulated malicious nodes are not identified by the global node, undesirable results would be obtained due to the Byzantine attacks \citep{lamport2019byzantine}. 
In the literature of machine learning, researchers made massive efforts to defend FL from the negative impacts of attacks.
A popular strategy is to design DC algorithms that are more robust to these attacks \citep{Krum2017, yin2018byzantine, guerraoui2018hidden, xie2019zeno, xie2020zeno++,li2020learning,fang2020local, cao2021fltrust,wu2020federated,cao2021provably,IEEE2021}.
In statistical literature for distributed learning, similar efforts have been made in pursuit of robust global estimators in the presence of data heterogeneity across different local nodes \citep{duan2022heterogeneity,tong2022distributed,xichenAOS24}.
\cite{zhang2024anomaly} provide a comprehensive review of recent progress in the field, with different attack and defense strategies categorized into several groups as shown in their Figure 2, and point out that robust aggregation methods, such as the Median method ~\citep{yin2018byzantine}, the Trim-mean~\citep{yin2018byzantine} and FLTrust~\citep{cao2021fltrust}, are generally effective for all types of attacks.


Although these robust DC algorithms can alleviate the negative impacts of malicious attacks, it's apparently more ideal if we could identify the malicious nodes and eliminate their negative impacts completely.
To identify malicious nodes, \citet{Krum2017} propose a defense referred to as Krum that treats local nodes whose update vector is too far away from the aggregated barycenter as malicious nodes and precludes them from the downstream aggregation. 
\citet{guerraoui2018hidden} propose Bulyan, a process that performs aggregation on subsets of node updates (by iteratively leaving each node out) to find a set of nodes with the most aligned updates given an aggregation rule. 
\citet{xie2019zeno} compute a \emph{stochastic descendant score} (SDS) based on the estimated descendant of the loss function and the magnitude of the update submitted to the global node, and only include a predefined number of nodes with the highest SDS in the aggregation.
On the other hand, \citet{chen2021zero} propose a zero-knowledge approach to detect and remove malicious nodes by solving a weighted clustering problem via the K-means algorithm. The resulting clusters update the model individually and accuracy against a validation set is checked. All nodes in a cluster with significant negative accuracy impact are rejected and removed from the aggregation step.

While existing methods for malicious node detection have explored many different ways to detect malicious nodes, they all share a common nature: the detection is based on direct analysis of the proposed updates of local nodes, which are gradients of model parameters in most cases.
However, it is usually the case that the obtained gradients are high-dimensional, and different dimensions of the gradient vectors remain quite different in the range of values and follow very different distributions. 
For example, our analysis for the Fashion-MNIST data in this study involves a neural network with 29,132 parameters, leading to a gradient vector of extremely high dimensionality from each local node.
A statistical summary of gradient vectors from real data applications reveals that different dimensions of these vectors have very different numerical scales ranging from $10^{-4}$ to $10^3$.
These phenomena make it very challenging to precisely detect malicious nodes directly based on the node updates, as a few dimensions often dominate the final result. 
Although the weighted clustering method proposed by \citet{chen2021zero} could partially avoid this problem by re-weighting different update dimensions, it is often not trivial to determine the weights in a principled way. 

In this paper, we propose to resolve these critical challenges from a novel perspective: instead of working on the node updates directly, we propose to extract information about malicious nodes indirectly by transforming the node updates from numeric gradient values to rankings. 
Compared to the original numeric gradient values, whose distributions are difficult to model, the rankings are much easier to handle both theoretically and practically.
Moreover, as rankings are scale-free, we no longer need to worry about the scale difference across different dimensions.
A highly efficient method called MANDERA is proposed to separate the malicious nodes from the benign ones based on the transformed rankings. $K$-means is used to cluster all the nodes into two groups based on the ranking vectors. 
We prove that MANDERA can achieve perfect malicious node detection for a wide range of Byzantine attacks under mild conditions, and thus provide reliable defense for federated learning. 
A collection of real data analyses confirms the advantages of MANDERA over existing methods. The code is available in GitHub: \url{https://github.com/Imathatguy/DataPoisoningFL-Rank}.

\section{Mathematical formulation of federated learning with Byzantine attacks}
\label{sec:FL}

Suppose a FL system with $n$ local nodes, as illustrated in Figure~\ref{fig:WF}, is established to train a global neural network, with $\vtheta=(\theta_1,\cdots,\theta_p)$ being the network parameters. 
Let $\vtheta^*$ be the current guess of $\vtheta$ announced by the central node, and $\mM \in \R^{n \times p}$ be the message matrix received by the central node from the $n$ local nodes given $\vtheta^*$.
Define $\mM_{i,:}$ and $\mM_{:,j}$ as the $i^{th}$ row and $j^{th}$ column of $\mM$, respectively.
Row vector $\mM_{i,:}$ represents the message vector from local node $i$.
After receiving the message matrix $\mM$, the central node updates $\vtheta^*$ by an aggregation function $\alpha$ below:
\begin{equation}
    \vtheta^*_{new}=\vtheta^* - \alpha(\mM,\bPsi),
\end{equation}
where $\vtheta^*_{new}$ denotes the updated guess of $\vtheta$, $\bPsi$ denotes control parameters of the aggregation function $\alpha$ which updates $\vtheta^*$ with an additive term based on $\mM$. 
In many cases, the aggregation function $\alpha$ takes the column average of the message matrix $\mM$ multiplied by a learning rate.

Assume that $n_m$ malicious nodes are manipulated by the attacker to conduct Byzantine attacks in the above FL system, with the other $n_b = n-n_m$ nodes are benign nodes.
Let $\mathcal{I}_m$ and $\mathcal{I}_b$ be the indices of the malicious nodes and benign nodes, respectively.
Assume a benign node $i\in\cI_b$ is associated with a local dataset of $N_i$ data points, and mini-batch stochastic gradient descent \citep{cotter2011better}, the most popular training method for deep-learning models, is used to train model parameters $\vtheta$.
Let $N^*$ be the batch size.
In an epoch of the iterative training process, the loss function to guide the training of parameter $\vtheta$ at a benign node $i$ has the generic formulation of 
$$L(\vtheta,\mD_i)=\frac{1}{N^*}\sum_{l=1}^{N^*}L(\vtheta,\mD_{i,l}),$$ where $\mD_{i,l}$ is the $l^{th}$ data point from node $i$ in the current epoch and $\mD_{i}=\{\mD_{i,l}\}_{1\leq l\leq N^*}$.
For example, for an image classification problem with $K$ categories of images where $\mD_{i,l}=(\vx_{i,l},y_{i,l})$ with $\vx_{i,l}$ being the image to be classified and $y_{i,l}$ being the corresponding image classification label, a typical loss function is the cross-entropy loss defined as below:
$$L(\vtheta,\mD_{i,l})=-\sum_{k=1}^K\log\rho_k(\vtheta,\vx_{i,l})\cdot\bbI(y_{i,l}=k),$$
where $\rho_k(\vtheta,\vx_{i,l})$ is the predictive probability for image $\vx_{i,l}$ belonging to image category $k$ under the neural network with parameter $\vtheta$.
In this situation, we would expect the following message vector from a benign node $i\in\cI_b$:
\begin{equation}\label{eq:M-matrix}
    \mM_{i,:} = \frac{\partial L(\vtheta,\mD_i)}{\partial \vtheta}\mid _{\vtheta = \vtheta^*}
    =\frac{1}{N^*} \sum_{l=1}^{N^*} \frac{\partial L(\vtheta,\mD_{i,l})}{\partial \vtheta}\mid _{\vtheta = \vtheta^*},
\end{equation}
Further define
\begin{equation}\label{eq:M-pixel}
\mM_{i,j} =\frac{1}{N^*} \sum_{l=1}^{N^*} \frac{\partial L(\vtheta,\mD_{i,l})}{\partial \theta_j}\mid _{\vtheta = \vtheta^*}
\end{equation}
as the $j^{th}$ element of $\mM_{i,:}$.
Concrete examples of FL systems of the above form can be found in Section~\ref{sec:experiments}.

While a benign node $i\in\cI_b$ calculates $\mM_{i,:}$ based on its local data, a malicious node $i\in\cI_m$ attacks the FL system by manipulating $\mM_{i,:}$ in some way.
Here, we consider four types of Byzantine attacks that have been widely studied in the machine learning literature  \citep{baruch2019little,wu2020federated} , namely \emph{Gaussian attacks} (GA), \emph{sign flipping attacks} (SF), \emph{zero gradient attacks} (ZG), and \emph{mean shift attacks} (MS).  These attacks are defined below.

\begin{definition} [Gaussian attack] \label{def:gaussiannoise}
In a Gaussian attack, the attacker manipulates the malicious nodes by specifying their message vectors with independent Gaussian random vectors, i.e., let $\mM_{i,:} \sim \mathcal{N}(\vm_{b,:}, \bSigma)$ for any $i\in\cI_m$, where the mean vector $\vm_{b,:}= \frac{1}{n_b} \sum_{k \in \mathcal{I}_b} \mM_{k,:}$ and the covariance matrix $\bSigma$ is determined by the attacker.
\end{definition}

\begin{definition} [Sign flipping attack with Zero gradient attack as a special case] \label{def:SF}
In a sign flipping attack, the attacker manipulates the malicious nodes by specifying $\mM_{i,:}=-r \vm_{b,:} $ for any $i\in\cI_m$ with some $r>0$.
Moreover, if the attacker specifies $r=\frac{n_b}{n_m}$, we have $\sum_{i=1}^n \mM_{i,:} = 0$ at each epoch, and the sign flipping attack becomes the zero gradient attack where the aggregated gradient across all nodes degenerates to a vector of zeros.
\end{definition}

\begin{definition} [Mean shift attack] \label{def:MS}
In a mean shift attack, the attacker manipulates the malicious nodes by specifying 
$\mM_{i,:}=\vm_{b,:}-z\cdot\bgamma_{b,:}$ for any $i\in\cI_m$,
where $\bgamma_{b,:}=(\bgamma_{b,1},\cdots,\bgamma_{b,p})$ with $\bgamma_{b,j}=\sqrt{\frac{1}{n_b}\sum_{i \in \cI_b} (\mM_{i,j} - \vm_{b,j})^2}$, and $z>0$ is a parameter specified by the attacker to control the attack strength, whose default value is 
$z = \Phi^{-1}\left(\frac{n-2}{2(n-n_m)}\right)$ according to \cite{baruch2019little},
with $\Phi^{-1}(\cdot)$ being the inverse cumulative distribution function of the standard Gaussian distribution.
\end{definition}

\section{Malicious node detection via statistical analysis of a ranking matrix of messages}
\label{sec:methods}

\subsection{Properties of the message matrix $\mM$}
Assume that the $N_i$ data points in a benign node $i\in\cI_b$ are \emph{independent and identically distributed} (IID) samples from a data distribution $\mathbb{D}$. Since mini-batch gradient descent \citep{cotter2011better} is the most popular training method for deep models, we use $N^*$ to denote the batch size for each epoch. 
Because $\mM_{i,j}$ is the mean of a collection of IID random variables for a given $\vtheta^*$, 
according to \emph{Kolmogorov’s strong law of large numbers} (KSLLN) and \emph{central limit theorem} (CLT), it's straightforward to show that $\mM_{i,j}$ would converge to a constant, with $\sqrt{N^*}\mM_{i,j}$ behaving like a Gaussian random variable, when $N^*$ is reasonably large. 
Lemma \ref{lemma:behavior.benign} below formally states the approximate normality of $\mM_{i,j}$, suggesting that the message matrix $\mM$ of benign nodes behaves like a Gaussian random matrix asymptotically, with elements in each column being independent random variables from a common Gaussian distribution.
\begin{lemma} \label{lemma:behavior.benign}
Assume that local data sets in all the benign nodes are IID samples from a data distribution $\mathbb{D}$.
For any $i\in\cI_b$ and $j\in\{1,\cdots,p\}$, denote 
\begin{equation}
\mu_j=\E \left(\frac{\partial L(\vtheta,\mD_{i,l})}{\partial \theta_j}\right)\quad \text{and}\quad \sigma^2_j=\Var\left(\frac{\partial L(\vtheta,\mD_{i,l})}{\partial \theta_j}\right) < \infty.
\end{equation}
With $N^* \rightarrow \infty$, we have $\mM_{i,j} \rightarrow\mu_j$ almost surely (a.s.), $\sqrt{N^*}\mM_{i,j} \rightarrow^{\hspace{-0.3cm}d}\ \mathcal{N}\left(\mu_j,\sigma_j^2\right)$, and thus $\mM_{:,j}\sim\cN(\mu_j{\bm 1}_n,\frac{\sigma^2_j}{N^*}\bI_n)$ approximately where ${\bm 1}_n$  and $\bI_n$ are the $n$-dimensional vector of ones and identity matrix, respectively.

\end{lemma}

\begin{figure}[tb]
    \centering
    \includegraphics[width=0.8\linewidth]{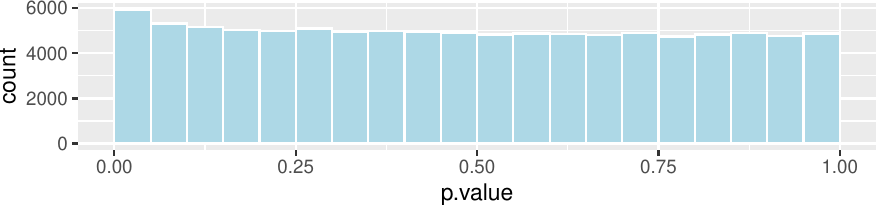}
    \caption{Histogram of $p$ -values of Pearson correlation tests for 100,000 pairs of columns randomly chosen from a message matrix $\mM$ generated from FASHION-MNIST data.}
    \label{fig:VerifyIndependenceAssumption}
\end{figure}

Further empirical studies show that columns of the message matrix of benign nodes are nearly independent of each other.
Let $\mM_{i,j}$ and $\mM_{i,k}$ be two elements of the message vector from a benign node $i\in\cI_b$, which come from the same row but different columns of matrix $\mM$. 
We find that $\mM_{i,j}$ and $\mM_{i,k}$ are nearly independent for all $(j,k)$ pairs, as the correlation between two model parameters $\vtheta_j$ and $\vtheta_k$ is often very weak in a large deep neural network with a huge number of parameters. 
To check the mutual independence of columns of $\mM$, we randomly chose 100,000 pairs of columns from a message matrix generated from an arbitrary steps of FL for the FASHION-MNIST data (see Section \ref{sec:experiments} for details), and conducted the Pearson correlation test for each pair.
Figure~\ref{fig:VerifyIndependenceAssumption} shows the distribution of $p$-values obtained from the 100,000 Pearson correlation tests, which  is very close to a uniform distribution, indicating that numbers in two different columns of $\mM$ are empirically uncorrelated. 
Given the fact that $\mM$ is a matrix of Gaussian random variables, the uncorrelated columns imply an independent random matrix immediately.
Due to these arguments, we treat the message matrix $\mM$ of benign nodes as a matrix of independent Gaussian random variables hereinafter for simplicity, with its rows following the multivariate Gaussian distribution below:
$$\mM_{i,:}\sim\cN(\bmu,\frac{1}{N^*}\bSigma_\bsigma),\ \forall\ i\in\cI_b,$$
where 
$$\bmu=(\mu_1,\cdots,\mu_p)',\ \bsigma=(\sigma_1^2,\cdots,\sigma_p^2)',\ \text{and}\ \bSigma_\bsigma=diag\{\sigma^2_1,\cdots,\sigma^2_p\}.$$

\subsection{From message matrix $\mM$ to rank matrix $\mR$}
Given a vector of real numbers $\va\in \R^{n\times 1}$, we define its ranking vector as $\vb=Rank(\va) \in \cP(1,\cdots,n)$, where the ranking operator $Rank$ maps the vector $a$ to an element in the permutation space $\cP(1,\cdots,n)$ containing all permutations of $(1,\cdots,n)$.
For example, $Rank(1.1, -2.0, 3.2) = (2,3,1)$ by assigning lower rankings to larger values. 
We adopt average ranking, when there are ties. 
For example, $Rank(1.1, 1.1, 3.2) = (2.5,2.5,1).$
With the \textit{Rank} operator, we can transfer a message matrix $\mM$ to a ranking matrix $\mR$ by replacing its column $\mM_{:,j}$ by the corresponding ranking vector $\mR_{:,j} = Rank(\mM_{:,j})$, i.e.,
\begin{equation}\label{eq:R-matrix}
    \mR=Rank_{col}(\mM)=(\mR_{:,1},\cdots,\mR_{:,p}).
\end{equation}
Apparently, the $(i,j)$ element of the rank matrix $\mR$ has the following formulation:
\begin{equation}\label{eq:R_ij}
    \mR_{i,j}=\sum_{k\neq i}\bbI(\mM_{k,j}>\mM_{i,j})+\frac{1}{2}\sum_{k\neq i}\bbI(\mM_{k,j}=\mM_{i,j})+1.
\end{equation}
For a fixed node $i$, and $1\leq j \neq k \leq p$, we observe that $R_{i,j}$ is a function of $M_{:,j}$ and $R_{i,k}$ is a function of $M_{:,k}$, which indicates that $R_{i,j}$ and $R_{i,k}$ are independent of each other when $M_{:,j}$ and $M_{:,k}$ are mutually independent.

Further, define 
\begin{equation}\label{eq:e-v}
e_i \triangleq \frac{1}{p} \sum_{j=1}^p \mR_{i,j} \qquad\text{and}\qquad v_i \triangleq \frac{1}{p} \sum_{j=1}^p \mR^2_{i,j} - e_i^2
\end{equation}
to be the mean and variance of $\mR_{i,:}$, the $i^{th}$ row of the rank matrix $\mR$.
As we will show in later subsections, $(e_i,v_i)$ can serve as an effective indicator to judge whether the node $i$ is a malicious node under a wide range of Byzantine attacks.

\subsection{Limiting behavior of $(e_i,v_i)$ for benign and malicious nodes}
Define 
\begin{equation}\label{eq:Limit_ei_vi}
    e^*_i\triangleq\lim_{p\rightarrow\infty}\frac{1}{p} \sum_{j=1}^p\bbE(\mR_{i,j})
    \quad\text{and}\quad 
    v^*_i\triangleq\lim_{p\rightarrow\infty}\frac{1}{p} \sum_{j=1}^p\bbE(\mR^2_{i,j})-(e_i^*)^2.
\end{equation}
 Given the fact that $\mR_{i,1},\cdots,\mR_{i,p}$ are independent random variables for a fixed node $i$, according to KSLLN we have 
$$\lim_{p\rightarrow\infty} e_i= e_i^*
\quad\text{and}\quad
\lim_{p\rightarrow\infty}v_i= v_i^*\quad a.s..$$

To get the concrete form of $e_i^*$ and $v_i^*$, we need to calculate $\bbE(\mR_{i,j})$ and $\bbE(\mR^2_{i,j})$ for $1\leq j\leq p$.
Lemma~\ref{lemma:behavior.benign} suggests that  $\mM_{i,j}\sim G_j=\cN(\mu_j,\sigma_j^2/N^*)$ for a benign node $i\in\cI_b$. 
According to the definitions of various Byzantine attacks in Section~\ref{sec:FL}, the message vectors of malicious nodes are IID samples from a different $p$-dimensional distribution $\bF$.
Let $F_j$ be the marginal distribution of $\bF$ at the $j^{th}$ dimension.
We have $\mM_{i,j}\sim F_j$ for a malicious node $i\in\cI_m$. 
Define 
\begin{eqnarray*}
r_{i,j}(x)&=&\bbE(\mR_{i,j}\mid\mM_{i,j}=x)=\bbE\Big(\sum_{k\neq i}\bbI(\mM_{k,j}>x)+\frac{1}{2}\sum_{k\neq i}\bbI(\mM_{k,j}=x)+1\Big)\\
&=&[n_b-\bbI(i\in\cI_b)][1- G_j(x)]+[n_m-\bbI(i\in\cI_m)][1-F_j(x)]\\
&&+\frac{n_m-\bbI(i\in\cI_m)}{2}[F_j(x)-F_j(x^-)] + 1,
\end{eqnarray*}
standing for the expected rank of $M_{i,j}$ among $\left\{M_{i,j}\right\}_{i=1}^{n}$ given $M_{i,j}=x$, where $G_j(x)$ and $F_j(x)$ are cumulated distribution functions of $G_j$ and $F_j$, respectively.
Apparently, for a fixed $j$, $r_{i,j}(x)$ may take only two distinct values, namely $r_{b,j}$ and $r_{m,j}$, depending on whether $i\in\cI_b$ or not, i.e.,
$$r_{i,j}(x)=r_{b,j}\cdot\bbI(i \in \cI_b)+r_{m,j}\cdot \bbI(i \in \cI_m),$$
where 
\begin{eqnarray*}
r_{b,j} &=& [n_b-1][1- G_j(x)]+n_m \cdot [1-F_j(x)] + 1,\\
r_{m,j} &=& n_b \cdot [1- G_j(x)]+[n_m -1] [1-F_j(x)] +\frac{n_m-\bbI(i\in\cI_m)}{2}[F_j(x)-F_j(x^-)] + 1.
\end{eqnarray*}
Term $\frac{n_m-\bbI(i\in\cI_m)}{2}[F_j(x)-F_j(x^-)] + 1$ in the above equation takes care of the tie rankings, in case that $F_j$ is a point mass distribution.

Similarly, define $$s_{i,j}(x)=\bbE(\mR^2_{i,j}\mid\mM_{i,j}=x)=\bbE\Big(\sum_{k\neq i}\bbI(\mM_{k,j}>x)+\frac{1}{2}\sum_{k\neq i}\bbI(\mM_{k,j}=x)+1\Big)^2.
$$ Apparently, for a fixed $j$, $s_{i,j}(x)$ may take only two distinct values, namely $s_{b,j}$ and $s_{m,j}$, depending on whether $i\in\cI_b$ or not, i.e.,
$$s_{i,j}(x)=s_{b,j}\cdot\bbI(i \in \cI_b)+s_{m,j}\cdot \bbI(i \in 
\cI_m),$$ 
where the concrete form of $s_{b,j}$ and $s_{m,j}$ can be found in the supplementary file.

It's straightforward to see that 
\begin{eqnarray}
    \label{eq:ER_ij}
    \bbE(\mR_{i,j})=\bbE(r_{i,j}(\mM_{i,j}))=E_{bj}\cdot \mathbb{I}(i \in \mathcal{I}_b)+E_{mj}\cdot \mathbb{I}(i \in \mathcal{I}_m),\\
    \label{eq:ER2_ij}
    \bbE(\mR^2_{i,j})=\bbE(s_{i,j}(\mM_{i,j}))=S_{bj}\cdot \mathbb{I}(i \in \mathcal{I}_b)+S_{mj}\cdot \mathbb{I}(i \in \mathcal{I}_m),
\end{eqnarray}
where
$$E_{bj}=\int_{-\infty}^{\infty}r_{b,j}(x) dG_j(x)\quad\text{and}\quad E_{mj}=\int_{-\infty}^{\infty}r_{m,j}(x) dF_j(x),$$
$$S_{bj}=\int_{-\infty}^{\infty}s_{b,j}(x) dG_j(x)\quad\text{and}\quad S_{mj}=\int_{-\infty}^{\infty}s_{m,j}(x) dF_j(x).$$

Thus, we have
\begin{eqnarray}
\label{eq:EbEm}
\lim_{p\rightarrow\infty}\frac{1}{p} \sum_{j=1}^p\bbE(\mR_{i,j})
&=&E_b\cdot\mathbb{I}(i \in \mathcal{I}_b)+ E_m\cdot\mathbb{I}(i \in \mathcal{I}_m),\\
\label{eq:VbVm}
\lim_{p\rightarrow\infty}\frac{1}{p} \sum_{j=1}^p\bbE(\mR^2_{i,j})
&=& S_b\cdot\mathbb{I}(i \in \mathcal{I}_b)+S_m\cdot\mathbb{I}(i \in \mathcal{I}_m),
\end{eqnarray}
where 
\begin{eqnarray*}
E_b=\lim_{p\rightarrow\infty}\frac{1}{p} \sum_{j=1}^p E_{bj},&& E_m=\lim_{p\rightarrow\infty}\frac{1}{p} \sum_{j=1}^p E_{mj},\\
S_b=\lim_{p\rightarrow\infty}\frac{1}{p} \sum_{j=1}^p S_{bj}, && S_m=\lim_{p\rightarrow\infty} \frac{1}{p} \sum_{j=1}^p S_{mj}.
\end{eqnarray*}
Further define $V_b=S_b-E_b^2$ and $V_m=S_m-E_m^2$.
Plugging in Eq. (\ref{eq:EbEm}) and (\ref{eq:VbVm}) into Eq. (\ref{eq:Limit_ei_vi}), we have
\begin{eqnarray*}
e^*_i
&=&\lim_{p\rightarrow\infty}\frac{1}{p} \sum_{j=1}^p\bbE(\mR_{i,j})
=E_b\cdot\mathbb{I}(i \in \mathcal{I}_b)+ E_m\cdot\mathbb{I}(i \in \mathcal{I}_m),\\
v^*_i
&=&\lim_{p\rightarrow\infty}\frac{1}{p} \sum_{j=1}^p\bbE(\mR^2_{i,j})-(e_i^*)^2
=V_b\cdot\mathbb{I}(i \in \mathcal{I}_b)+V_m\cdot\mathbb{I}(i \in \mathcal{I}_m).
\end{eqnarray*}

Summarizing the above theoretical analysis, we reach to the following theorem.
\begin{theorem} [Behavior under a general Byzantine attack] \label{thm:converge.general.byzantine}
For a FL system under a general Byzantine attack where the message vectors from benign nodes and malicious nodes follow two distinct $p$-dimensional distributions $G(\cdot)$ and $F(\cdot)$ respectively, we have
\begin{eqnarray*}
e_i&\rightarrow& E_b\cdot\mathbb{I}(i \in \mathcal{I}_b)+ E_m\cdot\mathbb{I}(i \in \mathcal{I}_m)\ a.s.,\\
v_i&\rightarrow&V_b\cdot\mathbb{I}(i \in \mathcal{I}_b)+V_m\cdot\mathbb{I}(i \in \mathcal{I}_m)\ a.s.,
\end{eqnarray*}
when both $N^*$ and $p$ go to infinity.
\end{theorem}

\subsection{More detailed results for specific Byzantine attacks}
Theorem \ref{thm:converge.general.byzantine} provide us the generic results about the limiting behavior of benign and malicious nodes in terms of $(e_i,v_i)$ under a general Byzantine attack, revealing that $(e_i,v_i)$ would converge to two different limits, one for benign nodes and one for malicious nodes. 
Here, we examine in details four types of Byzantine attacks that have been widely studied in the machine learning literature \citep{baruch2019little,wu2020federated}, namely \emph{Gaussian attacks} (GA), \emph{zero gradient attacks} (ZG), \emph{sign flipping attacks} (SF), and \emph{mean shift attacks} (MS).


Because $\mM_{i,:}\sim\cN(\bmu,\frac{1}{N^*}\bSigma_\bsigma)$ approximately for all $i\in\cI_b$ when $N^*$ is reasonable large according to Lemma~\ref{lemma:behavior.benign}, the independence of $\{\mM_{i,:}\}_{i\in\cI_b}$ implies that $\vm_{b,:}= \frac{1}{n_b} \sum_{i \in \mathcal{I}_b} \mM_{i,:}$
would converge to $\bmu$ when $N^*$ and $n_b$ are reasonably large. 
Based on this fact, $(E_b,V_b)$ and $(E_m,V_m)$, the limit value of feature vector $(e_i,v_i)$ for a benign or malicious node, can be calculated and compared, as summarized in 
Theorem \ref{thm:convergetomiddle}.
For simplicity of the main texts, we reserve all theoretical proofs of this and later theorems in the supplementary file.

\begin{theorem} [Behavior under Gaussian attacks] \label{thm:convergetomiddle}
Under a Gaussian attack  and $N^{*}\rightarrow\infty$, {$F_j=\mathcal{N}(\mu_j, \bSigma_{jj})$ and $G_j=\mathcal{N}(\mu_j, \sigma^2_j/N^*)$ for $1\leq j\leq p$}, where $\bSigma$ is determined by the attackers, and thus we have in general
$$E_b=E_m=\frac{n+1}{2} \quad \text{and}\quad V_b\neq V_m,$$
unless the attacker designs the covariance matrix $\bSigma$ of the attacking Gaussian distribution very carefully, e.g., let $\bSigma=\frac{1}{N^*}\bSigma_\bsigma$, to make $V_b=V_m$.
\end{theorem}


Under a sign flipping attack, $(E_b,V_b)$ and $(E_m,V_m)$ can be calculated and compared in a similar way.
Define $\rho=\lim_{p \rightarrow \infty}\frac{\sum_{j=1}^p\bbI(\mu_j>0)}{p}$.
Theorem \ref{thm:SignFlipping} shows that $(E_b,V_b)\neq(E_m,V_m)$ in general, unless $\rho$ satisfies some restrictive conditions.

\begin{theorem} [Behavior under sign flipping attacks] \label{thm:SignFlipping}
Under a sign flipping attack  and $N^{*}\rightarrow\infty$, $F_j$ is a point mass distribution at $-r \mu_j$ and $G_j=\mathcal{N}(\mu_j, \sigma^2_j/N^*)$ for $1\leq j\leq p$, and thus 
we have in general
$$(E_b,V_b)\neq(E_m,V_m),$$
unless $\rho$ equals to $\frac{1}{2}$ and is the root of a specific quadratic equation whose concrete form is provided in the supplementary file.
\end{theorem}


Similar results hold for mean shift attacks as summarized by Theorem \ref{thm:MeanShift}. 
\begin{theorem} [Behavior under mean shift attacks] \label{thm:MeanShift}
Under a mean shift attack and $N^{*}\rightarrow\infty$, {$F_j$ is a point mass distribution at $\mu_j-z\cdot\sigma_j$ and $G_j=\mathcal{N}(\mu_j, \sigma^2_j/N^*)$ for $1\leq j\leq p$}, and thus we have in general
$$(E_b,V_b)\neq(E_m,V_m).$$
\end{theorem}

\begin{figure}[tb]
    \centering
    \includegraphics[width=\linewidth]{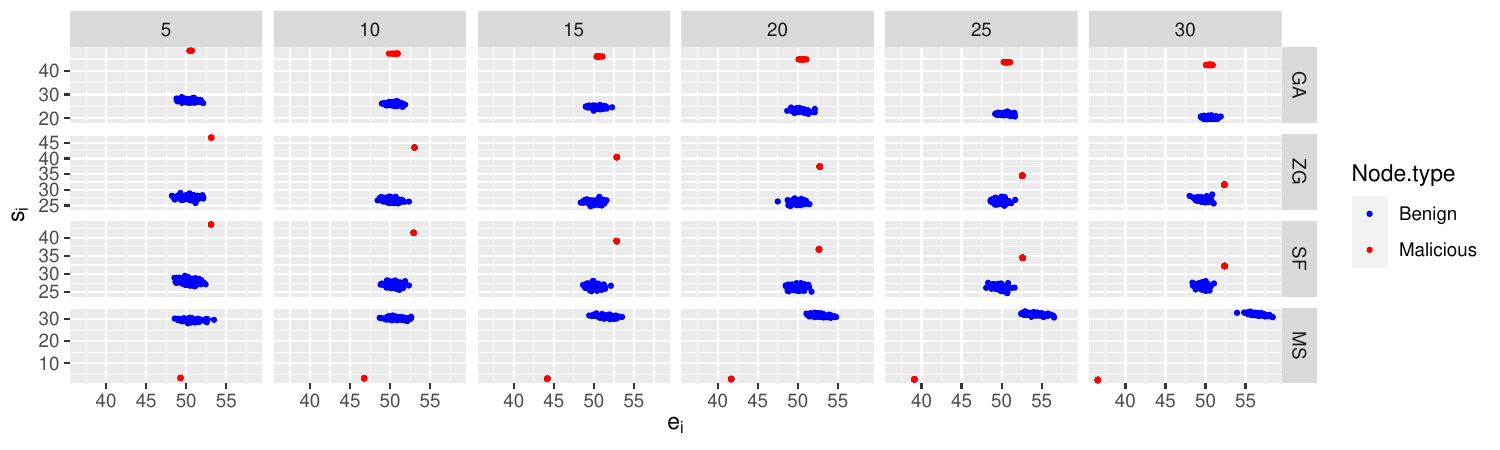}
    \caption{The scatter plots of $(e_i,s_i)$ for the 100 nodes under four types of attack as illustrative examples demonstrating ranking mean and standard deviation from the 1st epoch of training for the FASHION-MNIST dataset. Four attacks are Gaussian Attack (GA), Zero Gradient attack (ZG), Sign Flipping attack (SF) and Mean shift attack (MS).}
    \label{fig:both.metric.of.all.nodes}
\end{figure}

The above theoretical and practical analyses indicate that when $N^*$ and $p$ are both large, begin nodes and malicious nodes would form two clusters in the 2-dimensional space of $(e_i,v_i)$, because the chance to observe $(E_b,V_b)=(E_m,V_m)$ is usually close to zero.
Considering that the standard deviation $s_i=\sqrt{v_i}$ is typically of the similar scale of $e_i$,  hereinafter we employ $(e_i,s_i)$, instead of $(e_i,v_i)$, as the feature vector of node $i$ for malicious node detection.
Figure~\ref{fig:both.metric.of.all.nodes} shows the scatter plots of $(e_i,s_i)$ from the 1st epoch of training for the FL system for the {FASHION-MNIST} dataset ($N^*=10$ and $p=29132$) under four typical types of Byzantine attacks, with benign and malicious nodes highlighted in blue and red respectively.
It can be observed that malicious nodes and benign nodes are well separated into two clusters in all these scatter plots, indicating a proper clustering algorithm would be able to distinguish these two groups. 
We note that both $s_i$ and $e_i$ are necessary for malicious node detection, since in some cases (e.g., under Gaussian attacks) it is difficult to distinguish malicious nodes from benign ones based on $e_i$ only.

\subsection{Malicious node detection via node clustering based on $(e_i,s_i)$}

\begin{figure}[t]
    \centering
    \includegraphics[width=
    \linewidth]{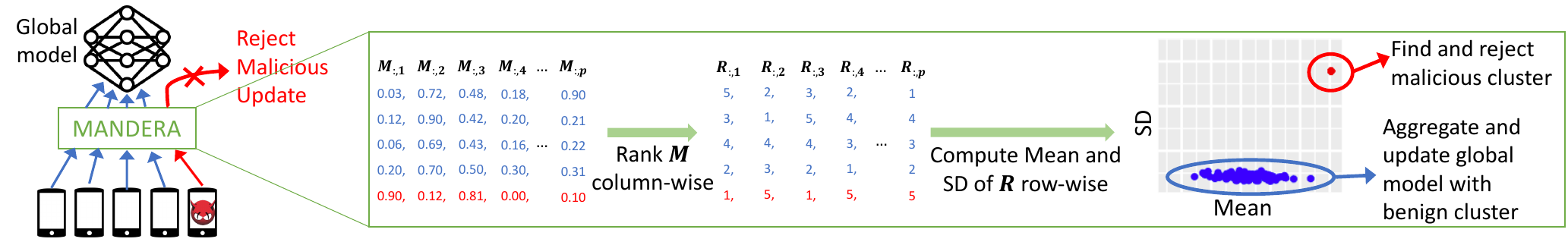}
    \caption{An overview of MANDERA.}
    \label{fig:mandera}
\end{figure}

Based on the above intuition, we propose \textit{MAlicious Node DEtection via RAnking} (MANDERA) to detect the malicious nodes, whose workflow is highlighted in Figure~\ref{fig:mandera} and detailed in Algorithm \ref{alg:MANDERA}. 
MANDERA can be applied to either a single epoch or multiple epochs. 
For the single-epoch mode, the input data $\mM$ is the message matrix received from a single epoch. 
For the multiple-epoch mode, the data $\mM$ is the column-concatenation of the message matrices from multiple epochs. 
Here, we focus on the single-epoch mode only, of which the multiple-epoch mode is a natural extension.

{We also test other clustering algorithms, such as hierarchical clustering and Gaussian mixture models \cite{fraley2002model}. 
It turns out that the performance of MANDERA is quite robust with different choices of clustering methods. 
Detailed results can be found in the supplementary file.}

\begin{algorithm}[ht]
\SetAlgoLined
\caption{MANDERA} 
\KwInput{The message matrix $\mM$.}
\begin{algorithmic} [1] \label{alg:MANDERA}
\STATE Convert the message matrix $\mM$ to the ranking matrix $\mR$ by applying \textit{Rank} operator;
\STATE Compute mean and standard deviation of rows in $\mR$, i.e., $\{(e_i,s_i)\}_{1\leq i\leq n}$;
\STATE Run the clustering algorithm $K$-means to $\{(e_i,s_i)\}_{1\leq i\leq n}$ with $K=2$ to discover the red and blue clusters, for example, highlighted in Figure~\ref{fig:mandera};
\STATE Given the assumption $n_b >\frac{n}{2}$, we claim the nodes in the larger cluster (denoted by $\hat\cI_b$) as the benign nodes, and the nodes in the smaller cluster (denoted by $\hat\cI_m$) as the malicious nodes.
\end{algorithmic}
\KwOutput{The predicted benign and malicious node sets $\hat\cI_b$ and $\hat\cI_m$.}
\end{algorithm}

The predicted benign nodes $\hat\cI_b$ obtained by MANDERA naturally leads to an aggregated message $\hat{\vm}_{b,:}=\frac{1}{\#(\hat\cI_b)} \sum_{i \in \hat\cI_b} \mM_{i,:}$.
Theorem \ref{thm:RobustnessGuarantee} shows that $\hat\cI_b$ and $\hat\vm_b$ lead to consistent estimations of $\cI_b$ and $\vm_{b,:} = \frac{1}{n_b} \sum_{i \in \cI_b} \mM_{i,:}$ respectively, indicating that MANDERA enjoys \emph{robustness guarantee} \citep{steinhardt2018robust} for Byzantine attacks.

\begin{theorem}[Robustness guarantee]\label{thm:RobustnessGuarantee}
For a FL system under Byzantine attacks with a fixed number of benign nodes and malicious nodes, assuming that the initiation of the $K$-means clustering described in Algorithm~\ref{alg:MANDERA} is appropriate to guarantee a successful clustering, 
we have:
\begin{eqnarray*}
\lim_{N^*, p\to \infty}\bbP(\hat\cI_b=\cI_b)=1, \ \lim_{N^*, p \rightarrow\infty} \bbE||\hat{\vm}_{b,:}-\vm_{b,:}||_2=0.
\end{eqnarray*}
\end{theorem}




{Based on this theorem, effective defense strategies again Byzantine attacks to FL can be established.
The most straightforward strategy is to run MANDERA at each epoch of the model training process when central node collects message vectors from all local nodes, and then conduct message aggregation for nodes in $\hat\cI_b$ only, with all suspected malicious nodes in $\hat\cI_m$ excluded.
In this study, we use this defense strategy by default.
The default defense strategy identifies malicious nodes at each epoch independently, which enjoys the flexibility in case that the attacker manipulates different nodes in different epochs.
If the attacker manipulates the same set of nodes in different epochs, an obviously more effective strategy is to claim benign and malicious nodes with joint consideration of the results in all $T$ epochs before the present epoch, namely $\{(\hat\cI_b^{(t)},\hat\cI_m^{(t)})\}_{1\leq t\leq T}$.
For example, we can define the following index $\alpha_i=\frac{1}{T}\sum_{t=1}^T\bbI(i\in\cI_m^{(t)})$ for node $i$ to quantify its ``prior probability'' to be a malicious node, and make decision for the current epoch with $\alpha_i$'s as additional reference.}

\section{Real data applications} \label{sec:experiments} 

\subsection{Federated learning of three real-world datasets under Byzantine attacks}
\begin{figure}[tb]
\centering
    \includegraphics[width=10cm]{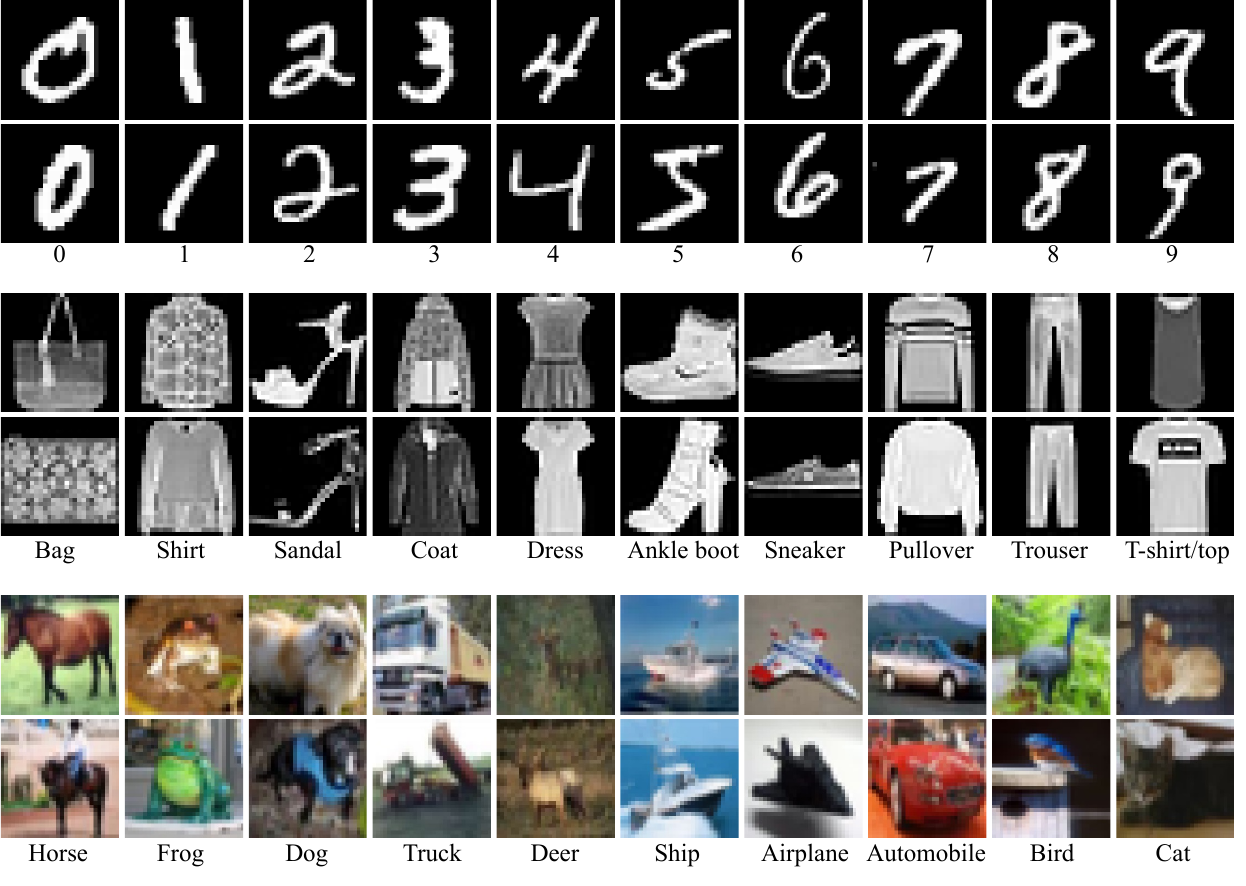}
    
    \caption{Typical images from MNIST, FASHION-MNIST and CIFAR-10 datasets. 
    The MNIST dataset is composed of 70,000 pictures of 10 classes of handwritten digits as showed in the top 2 rows. 
    The FASHION-MNIST dataset is composed of 70,000 pictures of 10 classes of apparels as showed in the middle 2 rows.
    The CIFAR-10 dataset is composed of 60,000 images of 10 classes of small objects as showed in the bottom 2 rows.
    }
    \label{fig:sampleimage}
\end{figure}

In this section, we adopted federated learning of three real-world image datasets, namely MNIST, FASHION-MNIST, and CIFAR-10 to evaluate the performance of MANDERA against existing methods. 
The MNIST~\citep{deng2012mnist} dataset is composed of 70,000 pictures of 10 classes of handwritten digits.
The FASHION-MNIST dataset~\citep{xiao2017online} is composed of 70,000 pictures of 10 classes of apparels.
The CIFAR-10 dataset~\citep{krizhevsky2009learning} is composed of 60,000 images of 10 classes of small objects.
Figure~\ref{fig:sampleimage} demonstrates typical images in these datasets.

For each dataset, a federated learning system was established to train a neural network for image classification, which reserved 1,000 holdout samples to evaluate the accuracy of the trained model, and randomly distributed all other samples to $n = 100$ local nodes connected to a global node as illustrated in Figure~\ref{fig:WF} (with each local node associated with one sub-datasets of the same size).
We train these neural networks for 25 epochs by a mini-batch SGD optimizer in the global node, with a batch size of 10, a learning rate of 0.01, and a momentum of 0.5. 
More details about the architecture of the involved neural networks can be found in the supplementary file. 

Given the above federated learning framework, we adopted the data poisoning framework in \cite{tolpegin2020data} and \cite{Byrd-saga-github} to conduct various Byzantine attacks, including GA, SF, ZG and MS.
In a typical attack, we randomly chose $n_m$ out of the 100 nodes as malicious nodes, where $n_m \in \{5, 10, 15, 20, 25, 30\}$.
We set $\bSigma=30\cdot\mI$ for Gaussian attacks with $\mI$ being the identity matrix, $r=3$ for the sign flipping attacks, and all the other attack settings to their default values.

Here, we compared MANDERA to 5 competing defense methods, including 3 {robust-estimation-based} defense methods, i.e., Median~\citep{yin2018byzantine}, Trim-mean~\cite{yin2018byzantine} and FLTrust~\cite{cao2021fltrust}, and 2 detection-based defense methods, i.e., Krum \citep{Krum2017} and Bulyan \citep{guerraoui2018hidden}.
These competing methods represent the state-of-the-art on defensing Byzantine attacks in FL. 
Because Krum and Bulyan require an assumed number of malicious nodes, we fed the true number of malicious nodes to them for simplicity, although such an action may lead to unfair comparison for MADERA due to information leaking.

We compared these competing methods from two angles: how well the 5 defense methods help suppress image mis-classification of the FL system due to malicious attacks, which is measured by the image classification accuracy of the final neural network obtained from the FL system with defense; and, how well MANDERA detects the malicious nodes in practice, which is measured by the detection accuracy of malicious nodes along the training process.
Additionally, we also tracked the trace plot of image classification accuracy and model log-loss along the model training process to highlight the dynamics of different defense methods.


\subsection{Defense performance in the IID scenario} \label{subsec:iid.results}

First, we evaluate and compare the defense performance of different methods when local data are IID samples from a data distribution, as assumed in Theorem~\ref{thm:converge.general.byzantine}. 
For this purpose, we randomly partitioned each of the three real datasets into $n=100$ equally sized sub-datasets, and assigned them to the 100 local nodes at random.
To guarantee a fair comparison, we conducted 10 parallel experiments for each experimental configuration and calculated the average performance of different defense strategies in these repeating experiments.
Please note that because malicious nodes were randomly chosen within each experiment, the set of malicious nodes are highly likely different across different experiments.


\begin{table}[tb]
\centering
\caption{Image classification accuracy of the global mode trained via FL after 25 epochs on the MNIST dataset with different defense strategies, under various Byzantine attacks with the number of malicious nodes $n_m$ ranging from 5 to 30 among 100 local nodes. The {\bf bold} highlights the best defense performance. The baseline accuracy with no attack is \textbf{98.45}.}
\resizebox{0.7\linewidth}{!}{  
\begin{tabular}{clcccccc}\toprule
Attack & Defense & $n_m=5$ & $n_m=10$ & $n_m=15$ & $n_m=20$ & $n_m=25$ & $n_m=30$ \\ \midrule
\multirow{6}{*}{GA} 
   & Median & 98.33 & 98.31 & 98.32 & 98.31 & 98.31 & 98.34 \\ 
   & Trim-mean & 98.45 & 98.43 & 98.41 & 98.38 & 98.38 & 98.35 \\ 
   & FLTrust & 95.33 & 65.22 & 61.02 & 37.45 & 11.37 & 12.17 \\ 
   & Krum & 96.77 & 96.63 & 96.78 & 96.89 & 96.90 & 96.90 \\ 
   & Bulyan & 98.46 & 98.43 & 98.40 & 98.36 & 98.35 & 98.29 \\ 
    & MANDERA & {\bf 98.48} & {\bf 98.46} & {\bf 98.44} & {\bf 98.43} & {\bf 98.44} & {\bf 98.42} \\ \midrule
      
  \multirow{6}{*}{ZG}  
   & Median & 98.17 & 98.00 & 97.74 & 97.36 & 96.77 & 96.10 \\ 
   & Trim-mean & 98.12 & 97.89 & 97.54 & 97.06 & 96.55 & 95.69 \\ 
   & FLTrust & 97.78 & 95.42 & 94.09 & 89.74 & 87.33 & 93.08 \\
   & Krum & 96.95 & 96.35 & 96.93 & 96.96 & 97.07 & 96.50 \\ 
   & Bulyan & 97.97 & 98.19 & 98.25 & 98.24 & 98.17 & 98.13 \\ 
   & MANDERA & {\bf 98.47} & {\bf 98.35} & {\bf 98.44} & {\bf 98.46} & {\bf 98.44} & {\bf 98.41} \\ \midrule
  \multirow{6}{*}{SF}   
   & Median & 98.16 & 98.00 & 97.75 & 97.33 & 96.78 & 96.14 \\ 
   & Trim-mean & 98.24 & 98.03 & 97.69 & 97.17 & 96.58 & 95.56 \\ 
   & FLTrust & 98.28 & 98.02 & 97.55 & 97.02 & 90.58 & 84.53 \\
   & Krum & 96.82 & 96.73 & 96.79 & 96.77 & 96.78 & 96.69 \\ 
   & Bulyan & 98.38 & 98.35 & 98.30 & 98.25 & 98.19 & 98.13 \\
   & MANDERA & {\bf 98.51} & {\bf 98.47} & {\bf 98.44} & {\bf 98.43} & {\bf 98.41} & {\bf 98.40} \\ \midrule
   
  \multirow{6}{*}{MS}  
   & Median & 98.41 & 98.39 & 98.33 & 98.28 & 98.25 & 98.23 \\ 
   & Trim-mean & 98.46 & 98.41 & 98.38 & 98.34 & 98.29 & 98.26 \\ 
   & FLTrust & 98.46 & 98.44 & 98.45 & 98.42 & 98.42 & 98.38 \\ 
   & Krum & 98.45 & 98.40 & 98.34 & 98.33 & 98.29 & 98.24 \\ 
   & Bulyan & 98.42 & 98.38 & 98.38 & 98.33 & 98.27 & 98.23 \\ 
   & MANDERA & {\bf 98.48} & {\bf 98.45} & {\bf 98.46} & {\bf 98.43} & {\bf 98.44} & {\bf 98.44} \\ 
   \bottomrule
\end{tabular}
}
\label{tab:MNIST.IamgeClassificationAccuracy}
\end{table}

\begin{figure}[tb]
\centering
    \includegraphics[width=10cm]{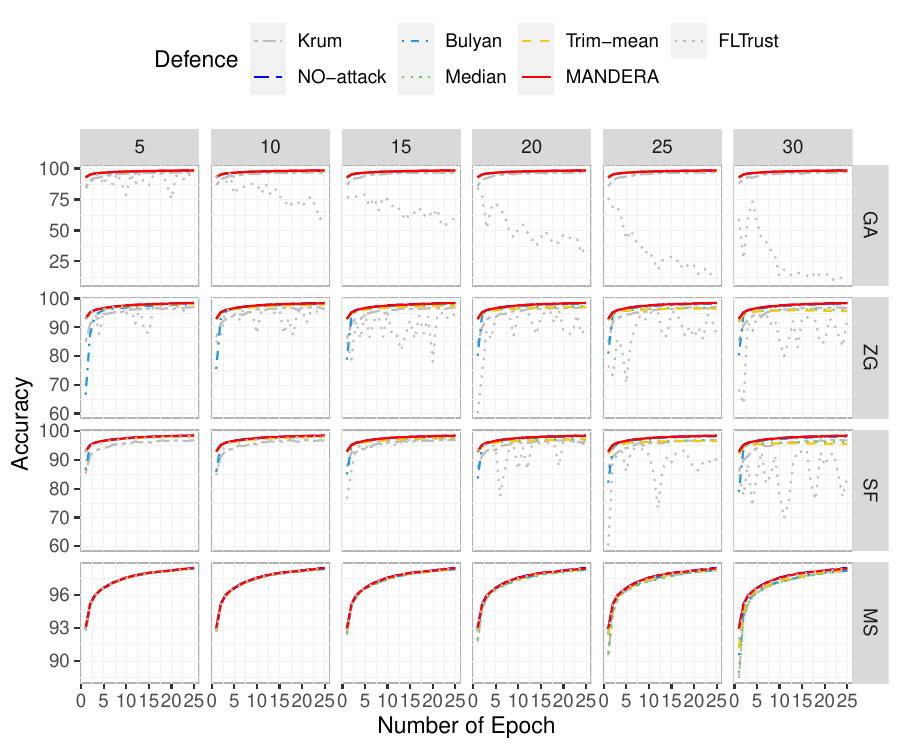}
    \includegraphics[width=10cm]{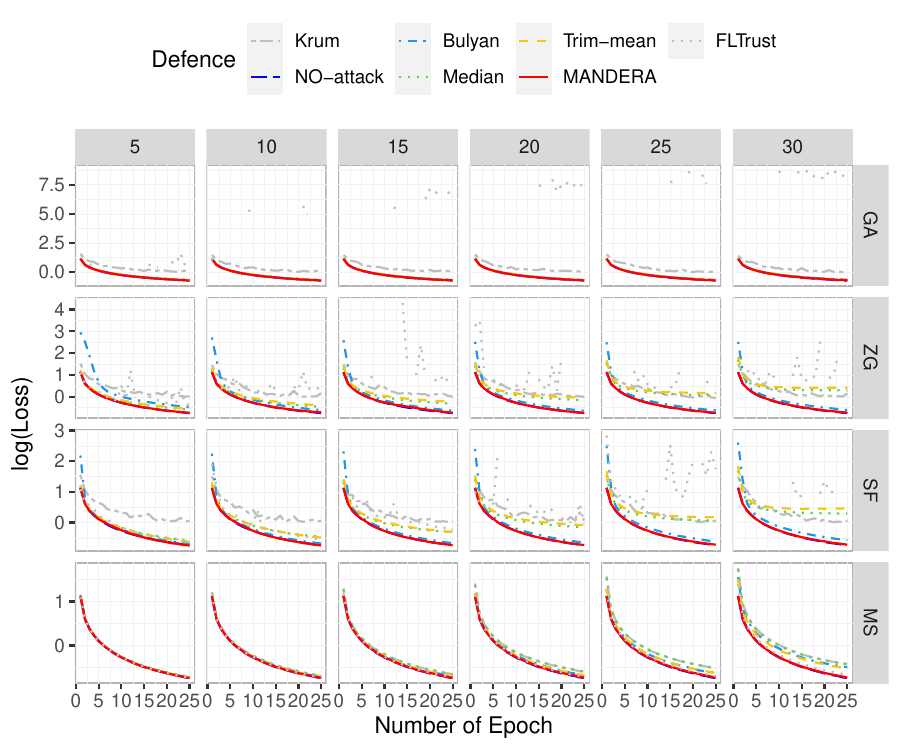}
    \caption{Trace plots of image classification accuracy and model log-loss of different defense methods under various types of Byzantine attacks on the MNIST dataset.
    }
    \label{fig:MNIST-TracePlot}
\end{figure}

Table \ref{tab:MNIST.IamgeClassificationAccuracy} demonstrates the image classification accuracy of the global model trained via FL after 25 epochs on the MNIST dataset, under four Byzantine attacks and six defense strategies with the number of malicious nodes $n_m$ ranging from 5 to 30. 
From the table, we can see the following facts.
First, achieving the highest final image classification accuracy in all cases, MANDERA universally outperforms all the other competing strategies on defensing a wide range of Byzantine attacks.
Second, MANDERA yields a robust image classification accuracy that is very close to the baseline accuracy of 98.45 when no attack is conducted in all cases, suggesting that the FL system defensed by MADERA loses little learning efficiency in presence of various Byzantine attacks.
The other defense strategies, however, typically suffer from a significant loss of learning efficiency, in terms of a lower image classification accuracy, with the increase of $n_m$, and unstable performance across different types of Byzantine attack.
Moreover, Figure~\ref{fig:MNIST-TracePlot} shows the trace plots of image classification accuracy and log-loss of the FL system on the MNIST dataset under different defense strategies against various types of Byzantine attacks with different specifications of $n_m$, confirming that MANDERA enjoys a steady advantage over competing methods along the entire training process of the FL system in all cases.

\begin{figure}[tb]
     \centering
    \includegraphics[width=11cm]{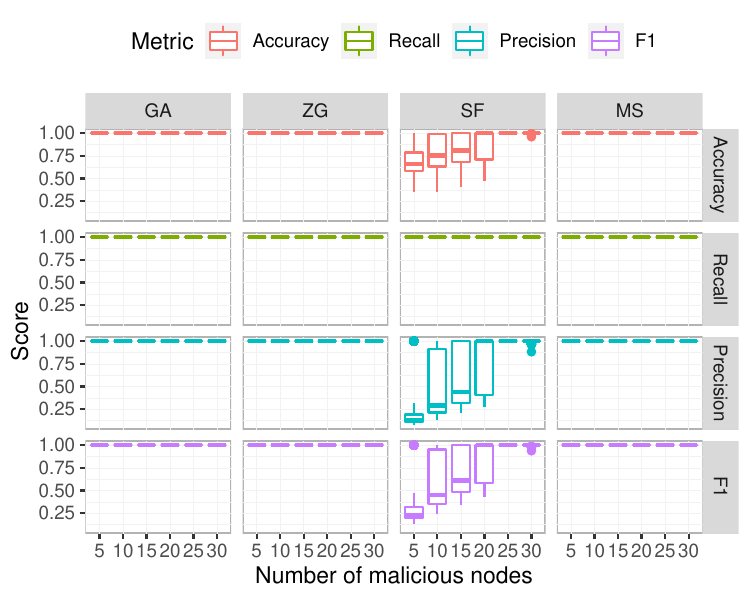}
    \caption{Performance of MANDERA on malicious node detection in the MNIST dataset under four types of Byzantine attack, with the number of malicious node increasing from 5 to 30. 
    } 
    \label{fig:MINST_DetectionAccuracy}
\end{figure}

To further evaluate the performance of MANDERA on malicious node detection in the MNIST dataset, we show in Figure~\ref{fig:MINST_DetectionAccuracy} the metrics of MANDERA on malicious node detection. The metrics are averaged over the 10 independent experiments.
From the figure, we can see that MANDERA perfectly detects all malicious nodes in every tested case with a high recall close to 1 all the time, and yields a high precision close to 1 for all cases expect for the SF attacks with small $n_m$.
Such results suggest that MANDERA is in general effective to distinguish malicious nodes and benign nodes, but may misclassify some benign nodes as malicious ones under SF attacks because SF sometimes is a weak attack depending on the value of $r$. When the model is nearly converged, the message vector $\mM_{k,:}$ of a benign node $k\in\cI_b$ is typically close to a $0$-vector, making the sign flipping attack a weak attack when $r=3$, and thus it's much more difficult to distinguish the malicious nodes and benign nodes.
Although mis-classification of some benign nodes as malicious ones is less ideal, such a bold action has little negative impact to the quality of the final neural network obtained by FL, because mis-classifying few benign nodes as malicious ones would only cause a little bit of loss of statistical efficiency when all malicious nodes have been successfully identified and excluded from the aggregation procedure of FL.
These results explain the excellent performance of MANDERA in defensing FL from various Byzantine attacks.

Parallel results for the other two datasets, i.e., FASHION-MNIST and CIFAR-10, are showed in the supplementary file. 
MANDERA yields very similar performance in these datasets as in the MNIST dataset, confirming that MANDERA is a widely effective defense strategy for FL.
Moreover, although MANDERA takes $K$-mean with $K=2$ as the default clustering method, we found that other clustering methods, e.g., hierarchical clustering and clustering based on the Gaussian mixture model, also work reasonably well under the framework of MANDERA
(see the supplementary file for the details).

\subsection{Defense performance in non-IID scenario}

In this section, we evaluate the defense performance of MANDERA when local nodes own heterogeneous datasets that deviate from the IID assumption.
For this purpose, we consider an additional application of MANDERA on the QMNIST dataset~\citep{qmnist-2019}, which reinforces the MNIST dataset~\citep{deng2012mnist} with extra identification information about the writer of each handwritten digit.
According to the identification information in the QMNIST dataset, 500 distinct writers were involved in generating the handwritten digits.
Here, we allocated the 70,000 handwritten digits in the dataset to the 100 local nodes in the FL system by writers, with approximately 5 writers for each local node, instead of purely randomly as in the previous subsection.
Because handwritten digits by a same writer are typically much more similar with each other than those by different writers,
such a setting is widely recognized as the non-IID setting in the FL community \citep{kairouz2021advances}. 
We kept all other configurations for analyzing the non-IID QMNIST dataset the same as in the IID scenario for the MNIST dataset in the previous subsection.

\begin{figure}[hb]
\centering
    \includegraphics[width=11cm]{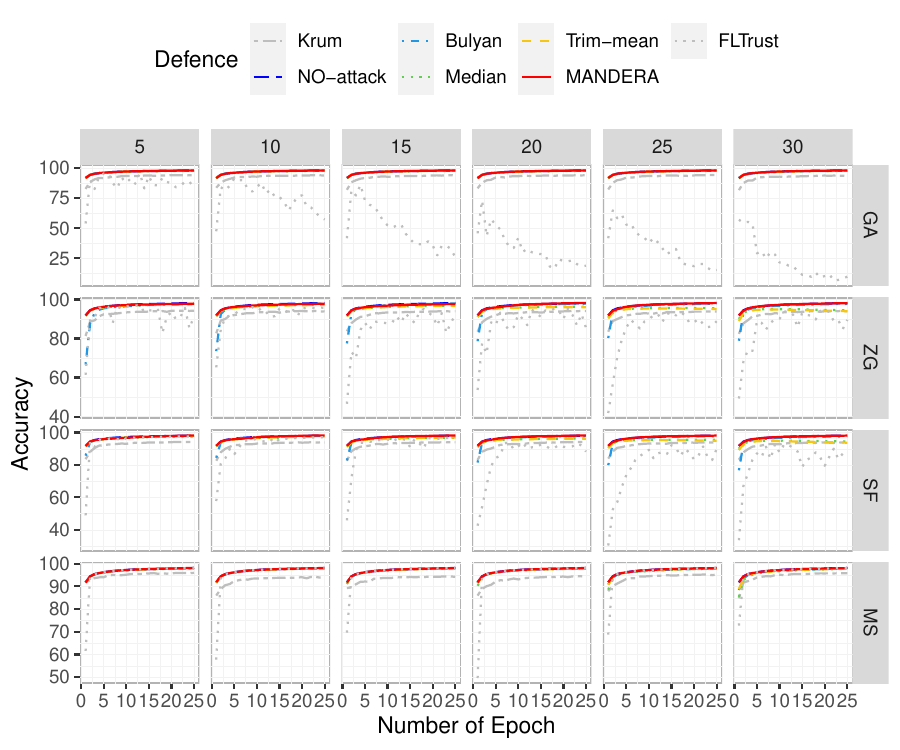}
    \includegraphics[width=11cm]{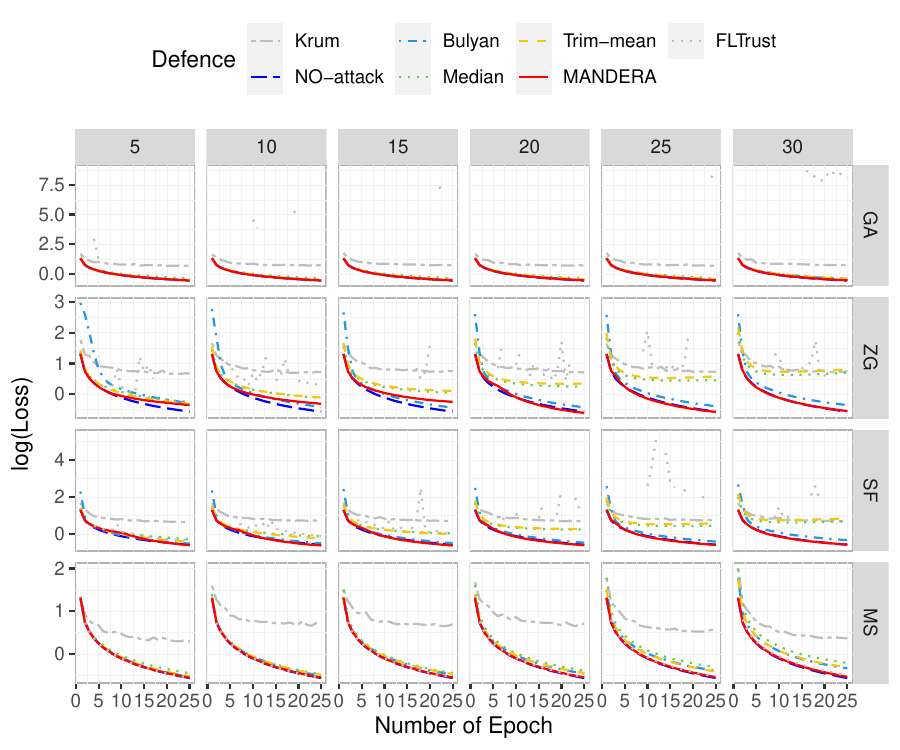}   
    \caption{Trace plots of image classification accuracy and model log-loss of different defense methods under various types of Byzantine attacks on the QMNIST dataset.}
    \label{fig:QMNIST-TracePlot}
\end{figure}

\begin{figure}[t]
\centering
    \includegraphics[width=0.80\linewidth]{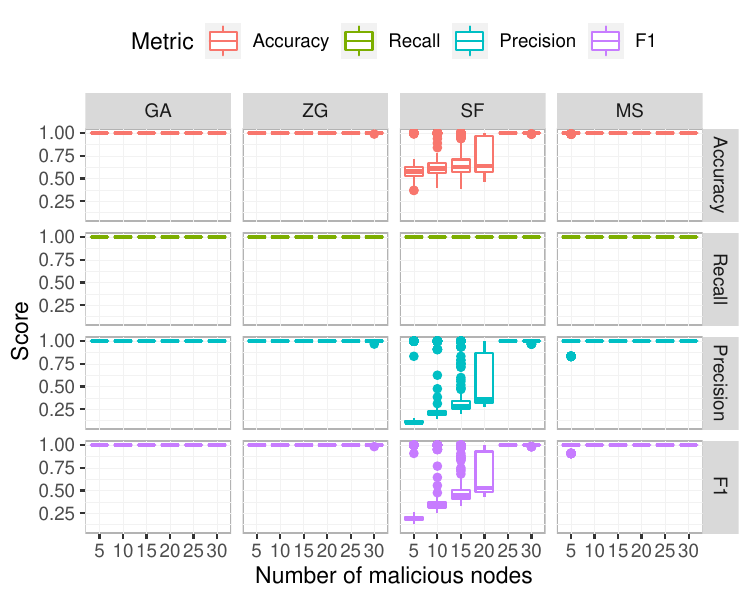}
    \caption{Malicious node detection by MANDERA for a Non-IID data set: QMNIST under four different Byzantine attacks, with the number of malicious node increasing from 5 to 30. }
    \label{fig:QMNIST-DetectionAccuracy}
\end{figure}

\begin{figure}
        \centering
        \begin{subfigure}[b]{0.32\textwidth} 
        \centering
        \includegraphics[width=\linewidth]{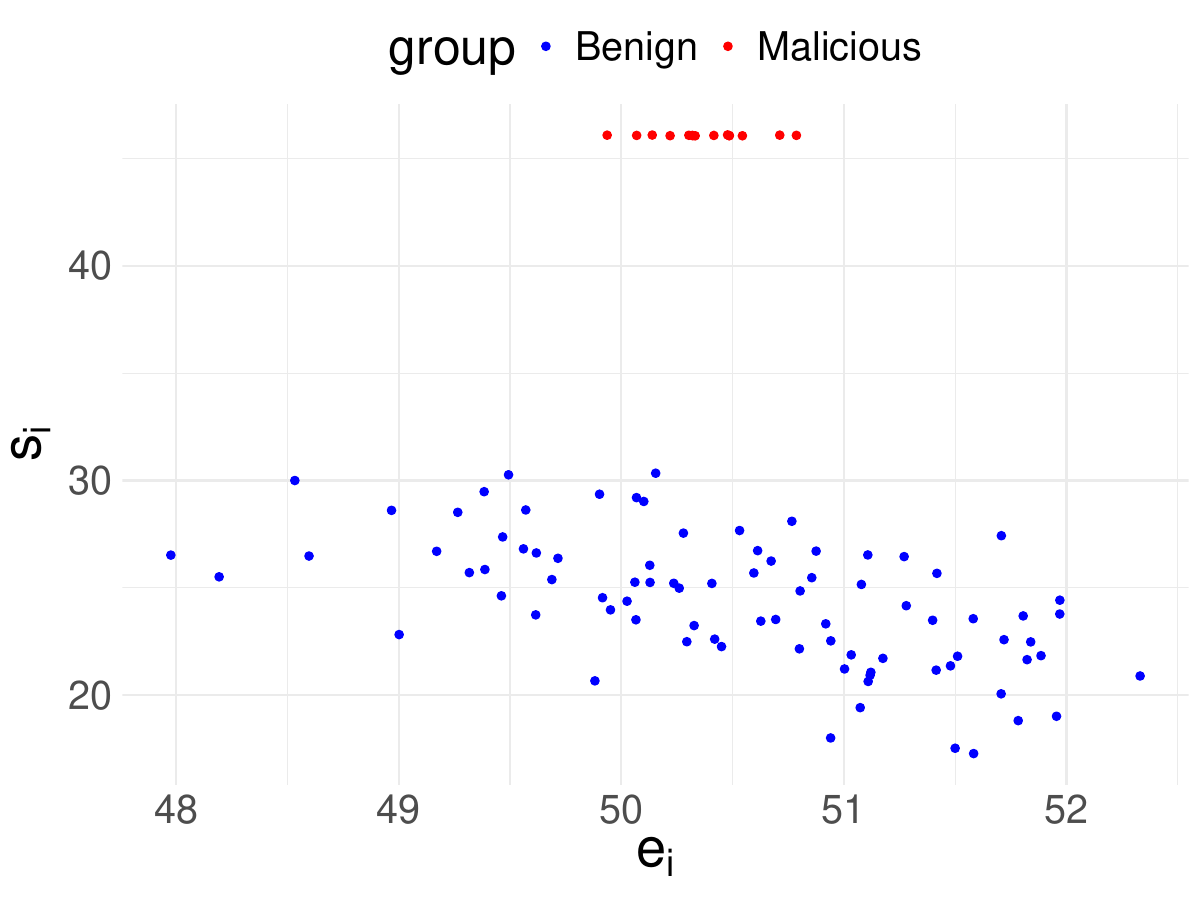}
        \caption{Scatter plot of $(e_i,s_i)$'s.}
        \label{fig:meanandsd}
    \end{subfigure}
\hfill
        \begin{subfigure}[b]{0.32\textwidth} 
        \centering
        \includegraphics[width=\linewidth]{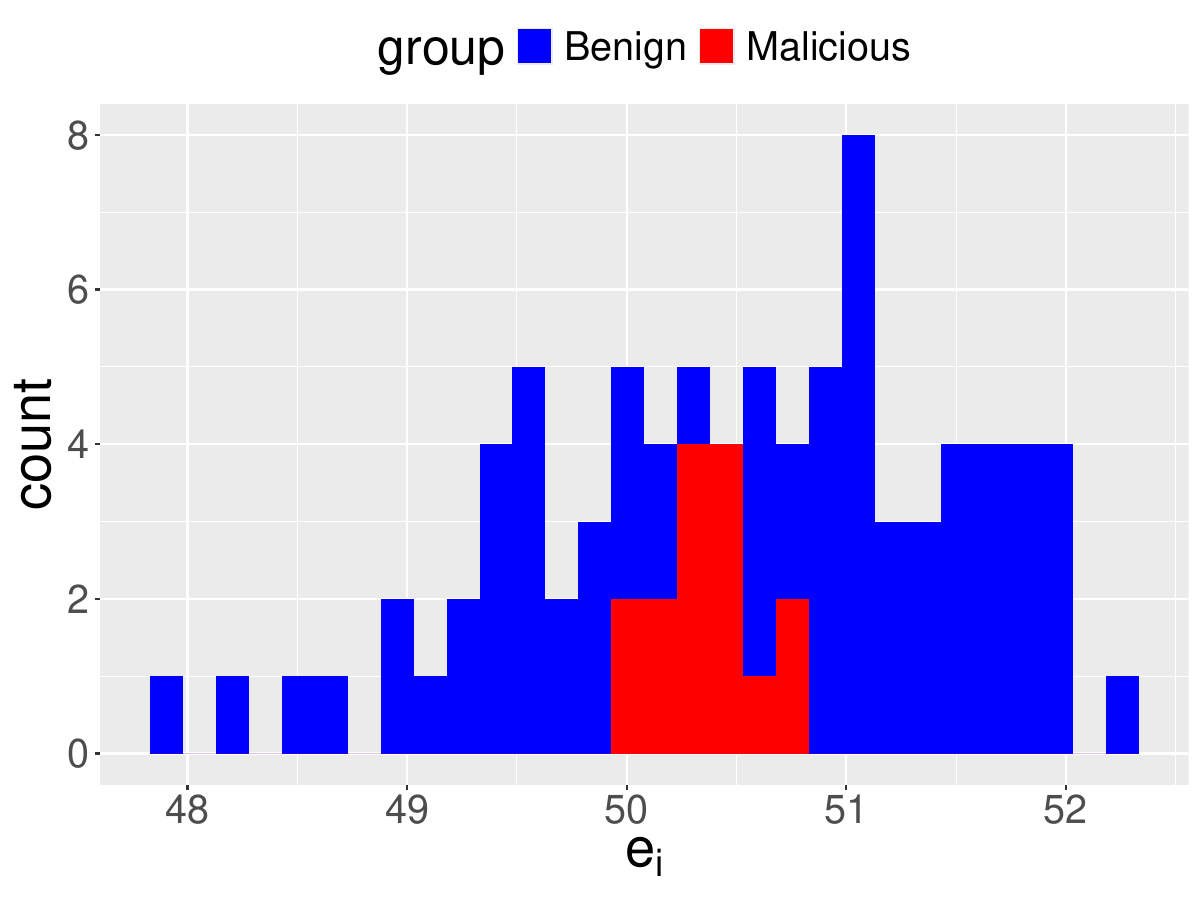}
        \caption{The histogram of $e_i$'s.}
        \label{fig:mean.histo}
    \end{subfigure}
        \hfill
        \begin{subfigure}[b]{0.32\textwidth} 
        \centering
       \includegraphics[width=\linewidth]{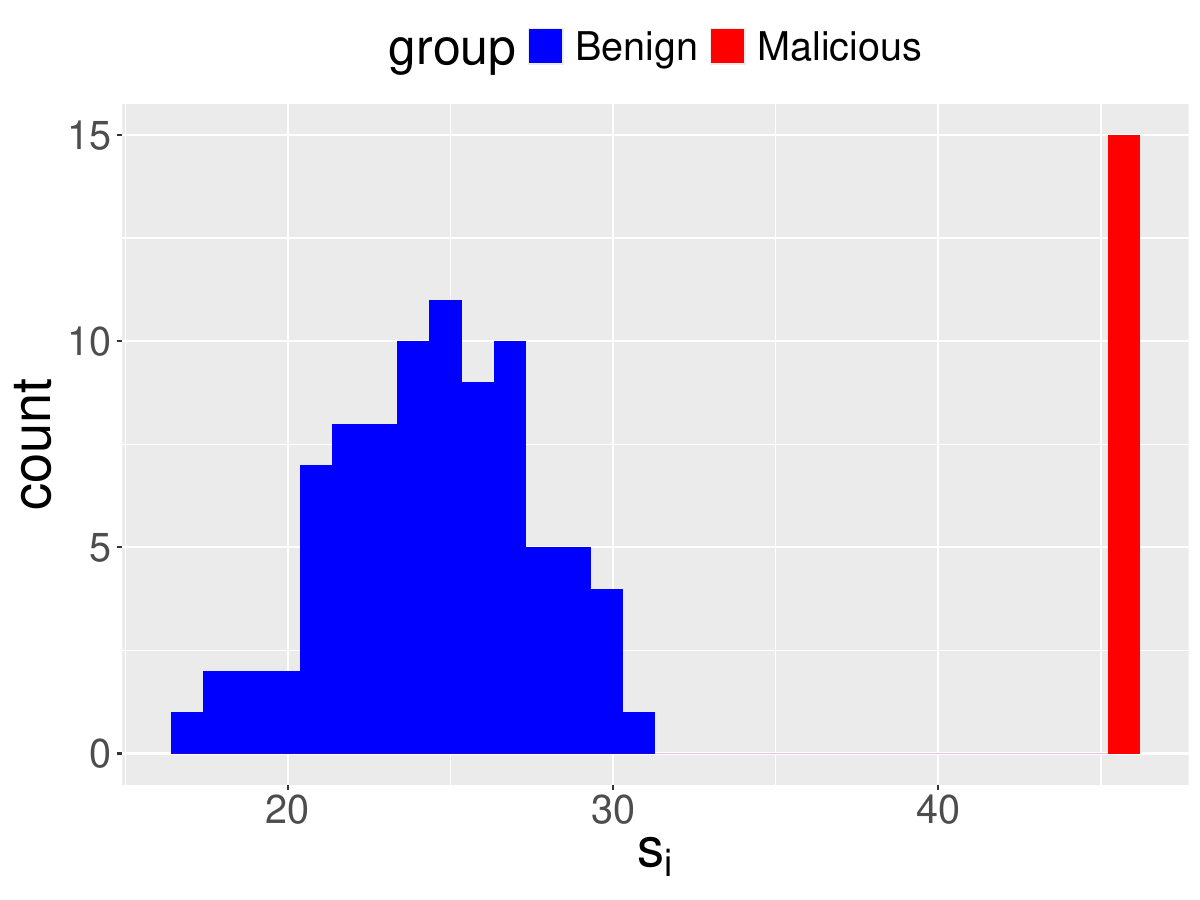}
        \caption{The histogram of $s_i$'s.}
        \label{fig:sd.histo}
    \end{subfigure}
    \caption{Joint and marginal distributions of $(e_i,s_i)$'s for malicious and benign nodes.}
    \label{fig:esPlot-nonIID}
\end{figure}

Parallel results for this non-IID scenario are showed in Table~\ref{tab:QMNIST.IamgeClassificationAccuracy}, Figure~\ref{fig:QMNIST-TracePlot} and \ref{fig:QMNIST-DetectionAccuracy}.
These results are very similar to the results from the MNIST dataset in the  IID scenario, showing that MANDERA is robust the deviation of the IID assumption.
{To investigate why and how MANDERA works robustly for the non-IID scenario, we demonstrate in Figure~\ref{fig:esPlot-nonIID} the joint and marginal distributions of $(e_i,s_i)$'s for the non-IID QMNIST data under Gaussian attack from 15 malicious nodes.
From the figures, we can clearly see that although the red cluster is more dispersed along the $X$-axis compared to its counterpart in Figure~\ref{fig:both.metric.of.all.nodes} due to the data heterogeneity of the non-IID scenario, the spatial separation of the red and blue clusters is perfectly kept.
This fact explains the robustness of MANDERA to heterogeneity in data across different nodes.
In other words, although data heterogeneity in the non-IID case may lead to an increase in $e_i$'s variability for malicious nodes, the ranking approach behind MANDERA still provides enough information to distinguish malicious nodes from benign ones, especially based on signals in $s_i$'s.}

\begin{table}[t]
\centering
\caption{Image classification accuracy of the global model trained via FL after 25 epochs on the QMNIST dataset with different defense strategies, under various Byzantine attacks with the number of malicious nodes $n_m$ ranging from 5 to 30 among 100 local nodes. The {\bf bold} highlights the best defense performance. The baseline accuracy with no attack is \textbf{98.12}.}
\resizebox{0.7\linewidth}{!}{  
\begin{tabular}{clcccccc}
  \toprule
Attack & Defense & $n_m=5$ & $n_m=10$ & $n_m=15$ & $n_m=20$ & $n_m=25$ & $n_m=30$ \\ \midrule
\multirow{6}{*}{GA} 
   & Median & 97.76 & 97.76 & 97.77 & 97.78 & 97.75 & 97.77 \\ 
   & Trim-mean & 98.08 & 98.04 & 98.00 & 97.96 & 97.91 & 97.85 \\ 
   & FLTrust & 83.48 & 57.32 & 25.75 & 18.80 & 15.43 & $\ $9.75 \\ \cline{2-8}
   & Krum & 94.16 & 93.87 & 93.95 & 94.10 & 94.27 & 93.89 \\ 
   & Bulyan & 98.09 & 98.07 & 98.06 & 98.02 & 97.99 & 97.88 \\ 
   & MANDERA & \textbf{98.11} & \textbf{98.11} & \textbf{98.12} & \textbf{98.10} & \textbf{98.10} & \textbf{98.08} \\ \midrule
   
  \multirow{6}{*}{ZG}  
   & Median & 97.59 & 97.27 & 96.84 & 96.33 & 95.54 & 94.45 \\ 
   & Trim-mean & 97.66 & 97.20 & 96.67 & 96.02 & 95.04 & 93.97 \\ 
   & FLTrust & 91.60 & 95.65 & 92.15 & 85.53 & 88.85 & 89.58 \\ \cline{2-8}
   & Krum & 94.21 & 93.90 & 93.92 & 94.11 & 93.84 & 93.95 \\ 
   & Bulyan & 97.58 & 97.83 & \textbf{97.90} & 97.87 & 97.79 & 97.71 \\
   & MANDERA & \textbf{97.85} & \textbf{97.78} & 97.64 & \textbf{98.21} & \textbf{98.13} & \textbf{98.09} \\ \midrule
   
  \multirow{6}{*}{SF} 
   & Median & 97.61 & 97.29 & 96.84 & 96.33 & 95.58 & 94.55 \\ 
   & Trim-mean & 97.82 & 97.52 & 96.97 & 96.21 & 94.98 & 93.75 \\ 
   & FLTrust & 97.75 & 97.21 & 96.65 & 88.25 & 89.99 & 88.29 \\ \cline{2-8}
   & Krum & 94.22 & 93.92 & 94.01 & 94.20 & 93.89 & 93.84 \\ 
   & Bulyan & 98.01 & 97.96 & 97.98 & 97.93 & 97.81 & 97.66 \\ 
   & MANDERA & \textbf{98.20} & \textbf{98.23} & \textbf{98.22} &\textbf{ 98.19} & \textbf{98.15} & \textbf{98.14} \\ \midrule
   
  \multirow{6}{*}{MS} 
   & Median & 97.88 & 97.96 & 97.96 & 97.90 & 97.79 & 97.70 \\ 
   & Trim-mean & 98.05 & 97.98 & 97.94 & 97.92 & 97.88 & 97.81 \\  
   & FLTrust & \textbf{98.13} & 98.11 & \textbf{98.12} & \textbf{98.10} & \textbf{98.09} & 98.06 \\ \cline{2-8}
   & Krum & 95.97 & 94.09 & 94.17 & 94.28 & 95.23 & 95.80 \\ 
   & Bulyan & 98.07 & 98.01 & 97.97 & 97.92 & 97.84 & 97.82 \\ 
   & MANDERA & 98.11 & \textbf{98.12} & 98.10 & 98.08 & 98.08 & \textbf{98.06} \\
   \bottomrule
\end{tabular}
}
\label{tab:QMNIST.IamgeClassificationAccuracy}
\end{table}


\subsection{Computational speed} \label{sec:computationaltime}

While yielding the best performance on defense various Byzantine attacks, MANDERA also enjoys fast computation because both ranking and clustering are computationally friendly operations. 
Table~\ref{tab:isolated_timing} shows the average running time of different defense strategies for conducting an aggregation stage in FL for the MNIST dataset, on one core of a Dual Xeon 14-core E5-2690 with a system RAM of 8 Gb and a single GPU of Nvidia Tesla P100. 
From the table we can see that MANDERA is the fastest among the detection-based defense strategies, although the robust-estimation-based defense strategies are in general much faster.


\begin{table}[tb]
\centering
\caption{Running time of different defense strategies for conducting an aggregation stage in FL for the MNIST dataset.
The running times are reported in the form of ``Mean ± Std" with millisecond as the time unit.
}
\resizebox{0.7\linewidth}{!}{  
\begin{tabular}{cccc}
\toprule
Detection-based defense      & Running time  & Robust-
estimation-based defense      & Running time  \\ \cmidrule(lr){1-2} \cmidrule(lr){3-4}
\em{\textbf{MANDERA}} & \ 643 \ \ \    ± \ \ \ 8.65 & Trimmed Mean         & 3.96  ± 0.41 \\
Krum& 1352 \ \   ± \  10.09 & Median               & 9.81  ± 3.88   \\
Bulyan               & 27209  ± \  233.4 & FLTrust     & \ 361 ± 4.07 \\  \bottomrule
\end{tabular}
}

\label{tab:isolated_timing}
\end{table}

\subsection{Robustness to hyperparameter specification and data partition}
{So far, the numerical results we reported are all based on experiments conducted under a particular specification of hyperparameters, such as learning rate and momentum, and a specific partition of data to different local nodes.
Careful researchers may naturally worry whether the superiority of MANDERA over other methods is robust to the specification of hyperparameters and data partition.

To investigate the impact of hyperparameter specification, we conducted a few additional experiments for each involved dataset, where hyperparameters like learning rate and momentum were specified to different values.
We found that MANDERA is not sensitive to the specification of hyperparameters, and typically achieves very good performance on malicious node detection even in the early stage of the model training process, where the global model is obviously far away from convergence.
This is because although hyperparameters have big impacts on the training process of a deep learning model, especially speed of convergence, they have little influence on MANDERA, because MANDERA does not rely on the convergence of global model to identify malicious nodes. 
Instead, MANDERA relies on the relative ranking of nodes to detect malicious nodes.
Such a phenomenon suggests that MANDERA is universally effective in detecting malicious nodes along all stages of the model training process, regardless of careful tuning of hyper-parameters. More detailed results can be found in the Supplementary.

To investigate the impact of data partition, we conducted some additional experiments on the MNIST dataset with different random data partitions.
We found that MANDERA maintains very stable performances across different data partitions, due to the relatively large sample size in each local node, indicating that MANDERA is robust to the dataset partition.
}

\section{Conclusions and Discussions} \label{sec:discussion.future}
We propose MANDERA, a highly efficient strategy to protect federated learning from Byzantine attacks that can detect malicious nodes effectively via statistical analysis of a ranking matrix obtained from the messages reported by decentralized devices.
Theoretical guarantee of MANDERA is established under mild conditions. 
A series of real data analyses for federated learning of large-scale image datasets via neural networks demonstrate the effectiveness and robustness of MANDERA in defensing a wide range of Byzantine attacks.

Compared to existing defense strategies based on malicious node detection, MANDERA enjoys a better and more robust defense performance, lower computation costs, and exemption of prior knowledge on the number of malicious nodes.
Compared to passive defenses via robust estimation such as Median and Trim-mean, which never try to detect the malicious nodes, MANDERA is more proactive and thus could put much more pressure on the attacker.
For FL systems equipped with MANDERA, conventional Byzantine attacks, such as GA, SF, ZG and MS, are not effective anymore, because malicious nodes manipulated by the attacker can be easily identified due to their obvious abnormal behaviors in the unique sight of MANDERA.
To reduce the chance to be captured, the attacker will have to reduce attack strength accordingly, which would significantly weaken the attack effect, or design more sophisticated attacks beyond regular Byzantine attacks.

In practice, more comprehensive defenses can be achieved based on extensions of MANDERA. 
For example, incorporating higher order moments of each row in the ranking matrix beyond mean and variance may enhance the defense performance of MANDERA for some types of attacks;
considering clustering results of nodes in multiple epochs jointly, instead of in one epoch, may further improve the accuracy in malicious node detection.
Moreover, integrating MANDERA and robust estimation may lead to more effective defenses for some attacks.




Although the asymptotic theory of MANDERA is established under the IID data assumption when both $N^*$ and $p$ go to infinity, practical performance of MANDERA is robust to deviations from these ideal settings.
The real data application of the QMNIST dataset has showed the robustness of MANDERA to deviations of the IID data assumption.
In all real data applications in the study, we set $N^*=10$, a very small number.
However, MANDERA still works very well in such an extreme case, suggesting that MANDERA is not sensitive to the specification of $N^*$.
{ Actually, even when $N^*$ is small, message vectors of all benign nodes still follow a same distribution, although not necessarily close enough to a Gaussian distribution anymore. 
In this case, the framework of MANDERA is still valid due to symmetry, although many detailed calculations in theorems for specific Byzantine attacks are not accurate any more. }
Moreover, because $N^*$ and $p$ incorporate with each other in the framework of MANDERA, instead of competing with each other, MANDERA would work well as long as both numbers are reasonably large.
Thus, we do not need to consider the relative ratio of $N^*$ and $p$ anymore when they go to infinity, as in classic statistical problems such as regression analysis.
We also note that MANDERA is not sensitive to specifications of hyperparameters, such as learning rate and momentum, and convergence of the deep learning model.

\section*{Acknowledgment}
This work is partially supported by the National
Key Research and Development Program of China (Grant No. 2023YFF0614702) and the National Natural Science Foundation of China (Grant 12371269). 
We thank Tongliang Liu for insightful discussions at the early stage of this paper.

\begin{supplement}
\stitle{Code availability}
\sdescription{The code is available in GitHub: \url{https://github.com/Imathatguy/DataPoisoningFL-Rank}.}
\end{supplement}
\begin{supplement}
\stitle{Proofs and more results}
\sdescription{Proofs for the Theorems and more detailed results can be found here.}
\end{supplement}
\FloatBarrier
\bibliography{ranking}
\bibliographystyle{imsart-nameyear}

\clearpage

\end{document}


\begin{frontmatter}
\title{Supplementary to ``MANDERA: Malicious Node Detection in Federated Learning via Ranking''}
\runtitle{Supplementary file}

\maketitle

\end{frontmatter}

\FloatBarrier

\section{Technical Proofs}\label{app:TechnicalProofs}

\subsection{Calculation of $s_{b,j}$ and $s_{m,j}$} \label{app:s_b_and_s_m}

Following the same logic as calculating $r_{b,j}$ and $r_{m,j}$, we have
\begin{eqnarray*}
    s_{b,j} &=&n_b (1-G(x))[1+(n_b-1)(1-G(x))]+ n_m (1-F(x))[1+(n_m-1)(1-F(x))] \\
        &&+ 2n_m (n_b-1)(1-F(x))(1-G(x)) + 2 (n_b-1)(1-G(x)) +2 (n_m-1)(1-F(x)),\\
    s_{m,j} &=& (n_b-1) (1-G(x))[1+(n_b-2)(1-G(x))] \\
        && +(n_m-1) (1-F(x))[1+(n_m-2)(1-F(x))] \\ 
        && + \frac{1}{4}(n_m-1)^2 (F(x)-F(x^-)) + 2n_b(n_m-1) (1-F(x))(1-G(x))\\ 
        && +2n_b(n_m-1) (1-G(x))(F(x)-F(x^-)) + 2n_b (1-G(x)) \\
        && +2n_m (1-F(x)) +2(n_m-1) (F(x)-F(x^-)). 
\end{eqnarray*}

\subsection{Proof of Theorem 3.3} \label{proof:convergetomiddle}


$ $\\ \\
According to Theorem 3.3, we need to calculate
$(E_{bj}, E_{mj}, S_{bj}, S_{mj})$ for $1\leq j\leq p$ under the Gaussian attack, which are determined by marginal distributions $F_j$ and $G_j$ jointly.
Because $\mM_{i,j}\rightarrow\mu_j$ a.s. when $N^*$ goes to infinity according to Lemma 3.1, it's straightforward to see that $\vm_{b,j}=\frac{1}{n_b}\sum_{i\in\cI_b}\mM_{i,j}\rightarrow\mu_j$ a.s. as well.
Thus, we have $F_j=\mathcal{N}(\mu_j, \bSigma_{jj})$ under the Gaussian attack, and $G_j=\mathcal{N}(\mu_j, \sigma^2_j/N^*)$ steadily.

On one hand, because $F_j$ and $G_j$ are both symmetric about $\mu_j$, it is straightforward to see that
$$E_{bj} = E_{mj} 
=\frac{n+1}{2},\ 1\leq j\leq p.$$
And, thus
$$E_b = \lim_{p\rightarrow\infty} \frac{1}{p} \sum_{j=1}^p E_{bj} = \frac{n+1}{2}
= \lim_{p\rightarrow\infty} \frac{1}{p}\sum_{j=1}^p E_{mj} = E_m.$$

On the other hand, however, we have $S_{bj}\neq S_{mj}$ in general, unless $\bSigma_{jj}=\sigma^2_j/N^*$ and thus $F_j=G_j$.
Because $S_{bj}$ and $S_{mj}$ are both complex functions of $\sigma^2_j$ and $\Sigma_{j,j}$, as well as $n_m$, $n_b$ and $N^*$, it's almost impossible to have 
$$S_b=\lim_{p\rightarrow\infty} \frac{1}{p} \sum_{j=1}^p S_{bj}=\lim_{p\rightarrow\infty} \frac{1}{p} \sum_{j=1}^p S_{mj}=S_m,$$
if the attacker specifies the covariance matrix $\bSigma$ utilized in Gaussian attack arbitrarily.
Therefore, in general we have 
$$V_b=S_b-E^2_n\neq S_m-E^2_m=V_m.$$


\subsection{Proof of Theorem 3.4} \label{proof:SignFlipping}

$ $\\ \\
According to Theorem 3.4, we need to calculate $(E_{bj}, E_{mj}, S_{bj}, S_{mj})$ for $1\leq j\leq p$ under the sign flipping attack.
Because we have showed that $G_j=\mathcal{N}(\mu_j, \sigma^2_j/N^*)$ steadily and $\vm_{b,j}\rightarrow\mu_j$ a.s. when $N^*$ goes to infinity, it's straightforward to see that $F_j$ is a point mass at $-r\mu_j$ under the sign flipping attack.

In case that $\mu_j=0$, $G_j=\mathcal{N}(0, \sigma^2_j/N^*)$ and $F_j$ becomes a point mass at 0, leading to 
$$E_{bj}=E_{mj}=\frac{n+1}{2}\quad\text{and}\quad  S_{bj}>S_{mj}.$$
In case that $\mu_j>0$, we have $-r\mu_j<\mu_j$, and thus $\bbP(\mR_{i,j}<\mR_{k,j})\approx 1$ for $\forall\ i\in\cI_b$ and $\forall\ k\in\cI_m$ when $N^*$ is reasonably large, leading to
$$E_{bj} =\frac{n_b+1}{2}<E_{mj}=\frac{n+n_b+1}{2}\quad\text{and}\quad  S_{bj}<S_{mj}.$$
In case that $\mu_j<0$, we have $-r\mu_j>\mu_j$, and thus $\bbP(\mR_{i,j}>\mR_{k,j})\approx 1$ for $\forall\ i\in\cI_b$ and $\forall\ k\in\cI_m$ when $N^*$ is reasonably large, leading to
$$E_{bj} =\frac{n+n_m+1}{2}>E_{mj}=\frac{n_m+1}{2}\quad\text{and}\quad  S_{bj}>S_{mj}.$$

Therefore, we have
\begin{eqnarray*}
 E_b&=&\lim_{p\rightarrow\infty} \frac{1}{p} \sum_{j=1}^p E_{bj} = \rho\cdot\frac{n_b+1}{2} + (1-\rho)\cdot\frac{n+n_m+1}{2},\\
 E_m&=&\lim_{p\rightarrow\infty} \frac{1}{p} \sum_{j=1}^p E_{mj} =\rho\cdot\frac{n+n_b+1}{2} + (1-\rho)\cdot \frac{n_m+1}{2},
\end{eqnarray*}
which satisfy $E_b=E_m$ if and only if $\rho=\frac{1}{2}$.
Moreover, to have 
$$V_b=S_b-E^2_n= S_m-E^2_m=V_m,$$
we need $\rho$ to be a root of the above equation, which is a quadratic equation of $\rho$ whose coefficients are complex functions of $r$, $\mu_j$'s, $\sigma^2_j$'s, as well as $n_m, n_b$, $N^*$ and $p$.
Apparently, it's very rare in practice that $\rho=\frac{1}{2}$ is the root of target equation by chance.
It completes the proof.

\subsection{Proof of Theorem 3.5} \label{app:MeanShift}

$ $\\

According to Theorem 3.5, we need to calculate $(E_{bj}, E_{mj}, S_{bj}, S_{mj})$ for $1\leq j\leq p$ under the mean shift attack.
Because we have showed that $G_j=\mathcal{N}(\mu_j, \sigma^2_j/N^*)$ steadily and $\vm_{b,j}\rightarrow\mu_j$ a.s. when $N^*$ goes to infinity, it's straightforward to see that $F_j$ is a point mass at {$\mu_j-z\cdot\sigma_j$} under the mean shift attack, where {$z = \Phi^{-1}\left(\frac{n-2}{2(n-n_m)}\right)$} by default according to \cite{baruch2019little}.

In this case, the concrete values of $(E_{b,j},S_{b,j})$ and $(E_{m,j},S_{m,j})$ depends on the relative order of $\mu_j$ and $\mu_j-z\cdot\sigma_j$, which is determined by the sign of $z$, in a similar way as in Theorem 3.4.
In case that $z=0$, 
$G_j=\mathcal{N}(\mu_j,\sigma^2_j/N^*)$ and $F_j$ becomes a point mass at $\mu_j$, leading to 
$$E_{bj}=E_{mj}=\frac{n+1}{2}\quad\text{and}\quad S_{bj}>S_{mj},\ 1\leq j\leq p.$$
In case that $z>0$, we have $\mu_j-z\cdot\sigma_j<\mu_j$, and thus $\bbP(R_{i,j}<R_{k,j})\approx 1$ for $\forall\ i\in\cI_b$ and $\forall\ k\in\cI_k$ when $N^*$ is reasonably large, leading to
$$E_{bj} =\frac{n_b+1}{2}<E_{mj}=\frac{n+n_b+1}{2}\quad\text{and}\quad S_{bj}<S_{mj},\ 1\leq j\leq p.$$
In case that $z<0$, we have $\mu_j-z\cdot\sigma_j>\mu_j$, and thus $\bbP(R_{i,j}>R_{k,j})\approx 1$ for $\forall\ i\in\cI_b$ and $\forall\ k\in\cI_k$ when $N^*$ is reasonably large, leading to
$$E_{bj} =\frac{n+n_m+1}{2}>E_{mj}=\frac{n_m+1}{2}\quad\text{and}\quad S_{bj}>S_{mj},\ 1\leq j\leq p.$$
No matter in which case, we have
$$E_b=\lim_{p\rightarrow\infty} \frac{1}{p} \sum_{j=1}^p E_{bj}\neq\lim_{p\rightarrow\infty} \frac{1}{p} \sum_{j=1}^p E_{mj}=E_m\ 
\text{or }
S_b=\lim_{p\rightarrow\infty} \frac{1}{p} \sum_{j=1}^p S_{bj}\neq\lim_{p\rightarrow\infty} \frac{1}{p} \sum_{j=1}^p S_{mj}=S_m,$$
which means that $(E_b,S_b)\neq(E_m,S_m)$.
It completes the proof.

\subsection{Proof of Theorem 3.6} \label{proof:RobustnessGuarantee}


According to Theorem 3.6, when both $N^*$ and $p$ are large enough, with probability 1 there exist $(e_b,v_b)$, $(e_m,v_m)$ and $\delta>0$ such that $||(e_b,v_b)-(e_m,v_m)||_2>\delta$, and
$$||(e_i,v_i)-(e_b,v_b)||_2\leq\frac{\delta}{2}\ \mbox{for}\ \forall\ i\in\cI_b\quad\ \mbox{and}\quad ||(e_i,v_i)-(e_m,v_m)||_2\leq\frac{\delta}{2}\ \mbox{for}\ \forall\ i\in\cI_m.$$
Therefore, with a reasonable clustering algorithm such as $K$-mean with $K=2$, we would expect $\hat{\cI}_b=\cI_b$ with probability 1.

Because we can always find a $\Delta>0$ such that $||\mM_{i,:}-\mM_{j,:}||_2\leq\Delta$ for any node pair $(i,j)$ in a fixed dataset with a finite number of nodes, and $\hat{\vm}_{b,:}=\vm_{b,:}$ when $\hat\cI_b=\cI_b$, we have 
$$\bbE||\hat{\vm}_{b,:}-\vm_{b,:}||_2\leq\Delta\cdot\bbP(\hat\cI_b\neq\cI_b),$$
and thus

$$\lim_{N^*\rightarrow\infty}\lim_{p\rightarrow\infty} \bbE||\hat{\vm}_{b,:}-\vm_{b,:}||_2=0.$$
It completes the proof.

\newpage
\section{More detailed results for real data applications}
\counterwithin{figure}{section}
\renewcommand{\thefigure}{S-\thesection.\arabic{figure}}

\counterwithin{table}{section}
\setcounter{table}{0}
\renewcommand{\thetable}{S-\thesection.\arabic{table}}
\subsection{Architecture of neural networks}\label{App:NeuralNetwork}

\begin{figure}[tb]
    \centering
    \includegraphics[width=0.5\linewidth]{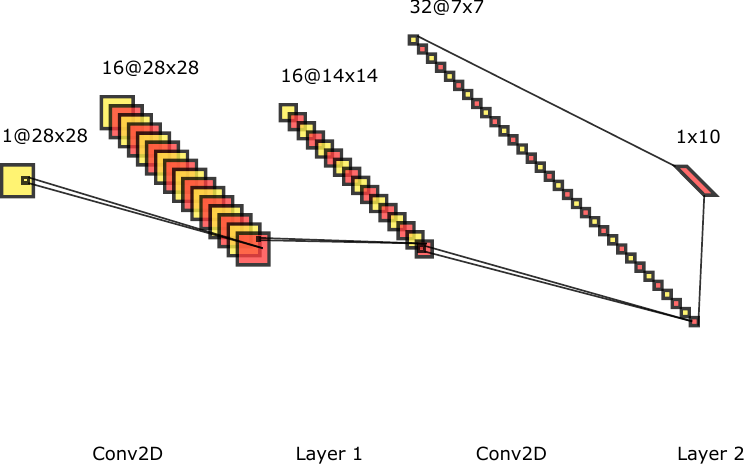}
    \caption{The neural network architecture for MNIST and FASHION-MNIST datasets.}
    \label{fig:mnistsnet}
\end{figure}

For MNIST and FASHION-MNIST datasets, we adopt the neural network architecture below with $p=29132$ parameters. 
The visualization can be found in Figure \ref{fig:mnistsnet}.
\begin{itemize}
\item Input Layer: image $3\times 28\times 28$, where $3$ is the number of channels of the image.
    \item Layer 1: 2D Convolution (kernel size is 5), Batch Normalization, ReLU Activation, Max pooling. The resulting dimension is $16 \times 14\times 14$, where 16 is the number of channels.
    \item Layer 2: 2D Convolution (kernel size is 5), Batch Normalization, ReLU Activation, Max pooling.  The resulting dimension is $32 \times 7\times 7$, where 32 is the number of channels.
    \item Output: $10$ Classes, Linear.
\end{itemize}

\begin{figure}[tb]
    \centering
    \includegraphics[width=0.5\linewidth]{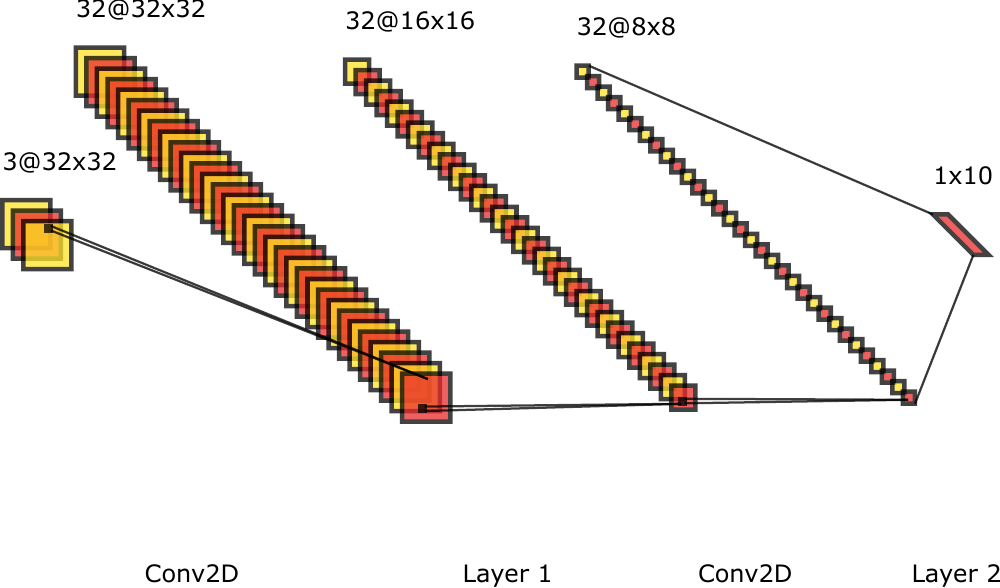}
    \caption{The neural network architecture for the CIFAR-10 dataset.}
    \label{fig:cifanet}
\end{figure}

For the CIFAR-10 dataset, we adopt the following neural network architecture with $p=30892$ parameters. 
The visualization can be found in Figure \ref{fig:cifanet}.
\begin{itemize}
\item Input Layer: image $3\times 32\times 32$, where $3$ is the number of channels of the image.
    \item Layer 1: 2D Convolution (kernel size is 3), Batch Normalization, ReLU Activation, Max pooling. The resulting dimension is $32 \times 16\times 16$, where 32 is the number of channels.
    \item Layer 2: 2D Convolution (kernel size is 3), Batch Normalization, ReLU Activation, Max pooling. The resulting dimension is $32 \times 8 \times 8$, where 32 is the number of channels.
    \item Output: $10$ Classes, Linear.
\end{itemize}

\newpage
\subsection{Results for the Fashion-MNIST dataset} \label{app:Fashion-MNIST}

\begin{figure}[hb]
    \centering
    \includegraphics[width=9cm]{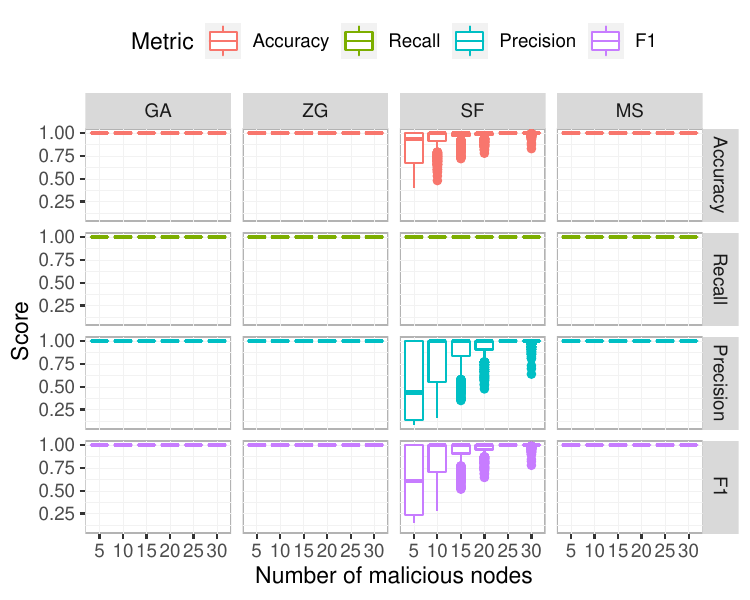}     
    \caption{The classification metrics of MANDERA on FASHION-MNIST dataset.}    
    \label{fig:fashion_DetectionAccuracy}
\end{figure}

\begin{figure}[ht]
    \centering
    \includegraphics[width=11cm]{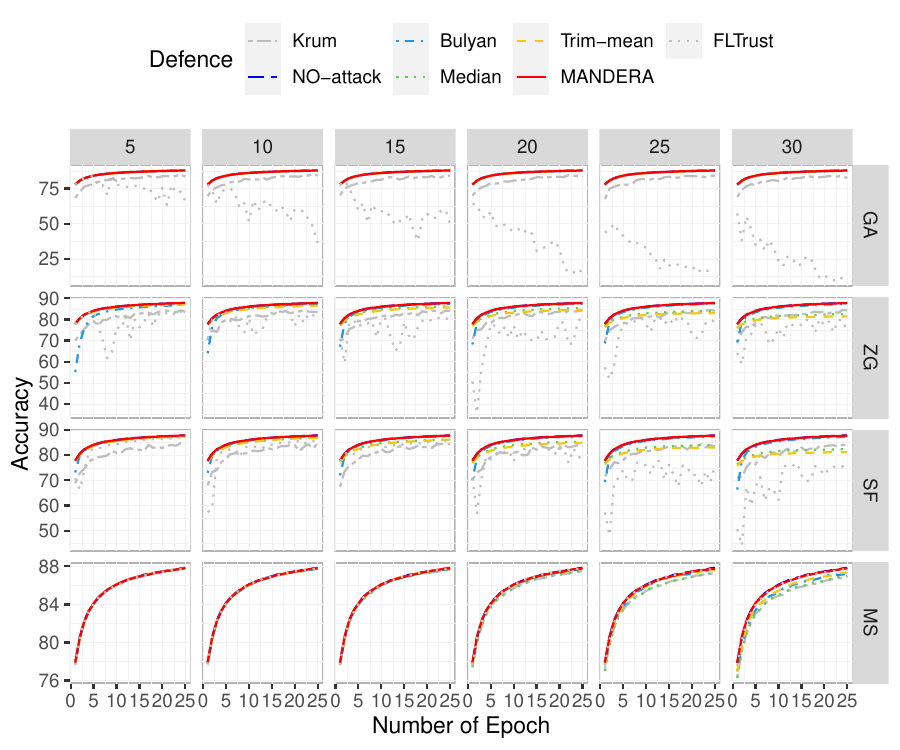}
    \includegraphics[width=11cm]{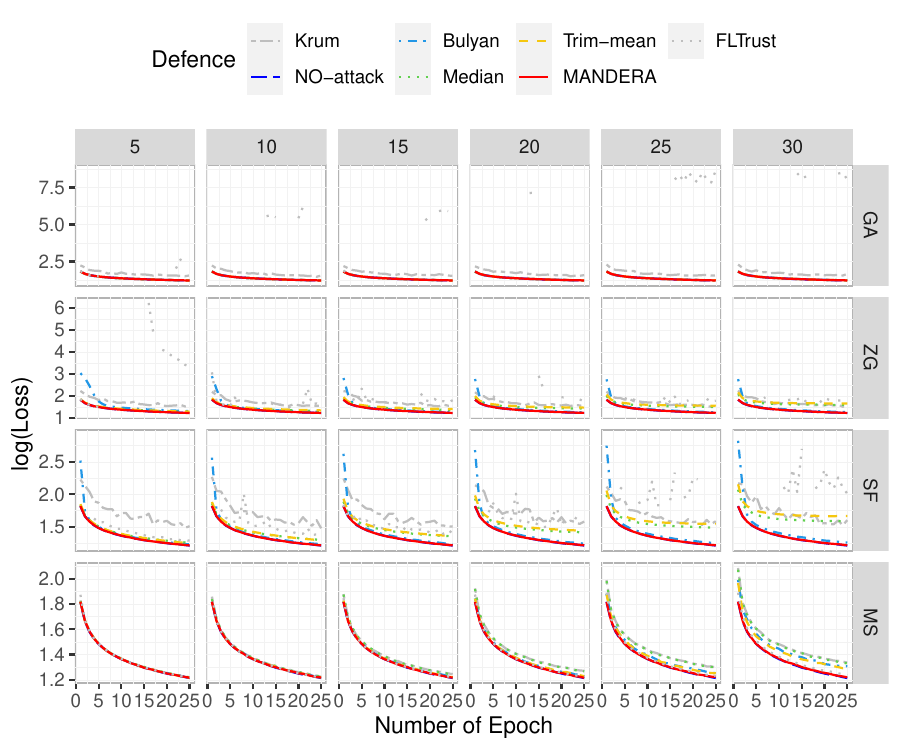}
    \caption{Trace plots of image classification accuracy and model log-loss of different defense methods on the FASHION-MNIST dataset.}
    \label{fig:FASHIONMNIST_TracePlot}
\end{figure}

Figure~\ref{fig:fashion_DetectionAccuracy} shows the metrics of MANDERA on the FASHION-MNIST dataset. 
Table~\ref{tab:FashionMNIST_Accuracy} shows the model accuracy at 25th epoch with $n_m$ denoting the number of malicious nodes among 100 nodes for FASHION-MNIST dataset.
 Figure~\ref{fig:FASHIONMNIST_TracePlot} shows the trace plot of accuracy and loss of different defense methods on the FASHION-MNIST dataset.

\begin{table}[tb]
\centering
\caption{FASHION-MNIST model accuracy at 25th epoch with $n_m$ denoting the number of malicious nodes among 100 nodes. The {\bf bold} highlights the best defense strategy under attack. The baseline accuracy when no attack is conducted is \textbf{87.83}.}
\label{tab:FashionMNIST_Accuracy}
\resizebox{0.8\linewidth}{!}{  
\begin{tabular}{clcccccc}
  \toprule
Attack & Defense & $n_m=5$ & $n_m=10$ & $n_m=15$ & $n_m=20$ & $n_m=25$ & $n_m=30$ \\ \midrule
  \multirow{6}{*}{GA} 
   & Median & 87.73 & 87.76 & 87.73 & 87.70 & 87.72 & 87.70 \\ 
   & Trim-mean & \textbf{87.85} & 87.78 & 87.75 & 87.74 & 87.72 & 87.73 \\ 
   & FLTrust & 66.13 & 36.35 & 50.20 & 17.85 & 16.00 & $\ $9.66 \\ \cline{2-8}
   & Krum & 83.66 & 84.13 & 84.09 & 83.30 & 84.22 & 82.32 \\ 
   & Bulyan & 87.80 & 87.80 & 87.79 & 87.73 & 87.67 & 87.69 \\ 
   & MANDERA & {87.81} & \textbf{87.83} & \textbf{87.82 }& \textbf{87.77} & \textbf{87.80} & \textbf{87.76} \\
   \midrule
  
  \multirow{6}{*}{ZG} 
   & Median & 87.36 & 86.91 & 86.20 & 85.33 & 84.07 & 82.45 \\ 
   & Trim-mean & 87.13 & 86.57 & 85.67 & 84.61 & 83.06 & 81.48 \\ 
   & FLTrust & 81.59 & 83.58 & 79.41 & 80.62 & 79.00 & 74.01 \\ \cline{2-8}
   & Krum & 83.56 & 83.57 & 84.11 & 84.33 & 84.10 & 84.30 \\ 
   & Bulyan & 86.88 & 87.38 & 87.49 & 87.45 & 87.48 & 87.38 \\ 
   & MANDERA & \textbf{87.79} & \textbf{87.81} &\textbf{ 87.84} & \textbf{87.72} & \textbf{87.76} & \textbf{87.78} \\ 
   \midrule
   
  \multirow{6}{*}{SF} 
   & Median & 87.40 & 86.91 & 86.21 & 85.36 & 84.11 & 82.31 \\ 
   & Trim-mean & 87.48 & 86.97 & 86.20 & 84.92 & 83.08 & 81.20 \\ 
   & FLTrust & 86.96 & 85.97 & 84.55 & 76.92 & 75.72 & 76.90 \\ \cline{2-8}
   & Krum & 84.49 & 84.71 & 84.43 & 83.58 & 83.61 & 83.72 \\ 
   & Bulyan & 87.60 & 87.64 & 87.62 & 87.50 & 87.47 & 87.35 \\ 
   & MANDERA & \textbf{87.85} & \textbf{87.79} & \textbf{87.82} & \textbf{87.79} & \textbf{87.77} & \textbf{87.74} \\ 
   \midrule
   
  \multirow{6}{*}{MS} 
   & Median & 87.75 & 87.78 & 87.69 & 87.52 & 87.26 & 86.99 \\ 
   & Trim-mean & 87.81 & \textbf{87.79} & 87.76 & 87.73 & 87.61 & 87.33 \\ 
   & FLTrust & 87.77 & 87.75 & 87.78 & 87.77 & \textbf{87.73} & 87.73 \\ \cline{2-8}
   & Krum & 87.82 & 87.77 & 87.66 & 87.50 & 87.36 & 86.89 \\ 
   & Bulyan & 87.81 & 87.78 & 87.75 & 87.75 & 87.60 & 87.21 \\ 
   & MANDERA & \textbf{87.81} & {87.78} & \textbf{87.78} & \textbf{87.79} & {87.71} & \textbf{87.79} \\ 
   \bottomrule
\end{tabular}
}
\end{table}

\newpage

\subsection{Results for the CIFAR-10 dataset} \label{app:CIFAR-10}

\begin{figure}[hb]
    \centering
    \includegraphics[width=9cm]{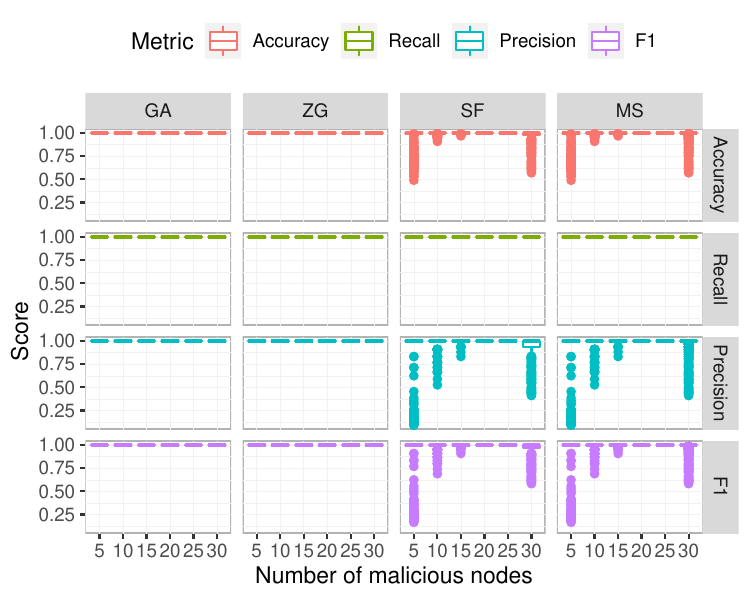}
    \caption{The classification metrics of MANDERA on CIFAR-10.}
    \label{fig:CIFAR_DetectionAccuracy}
\end{figure}

\begin{figure}[hb]
    \centering
    \includegraphics[width=11cm]{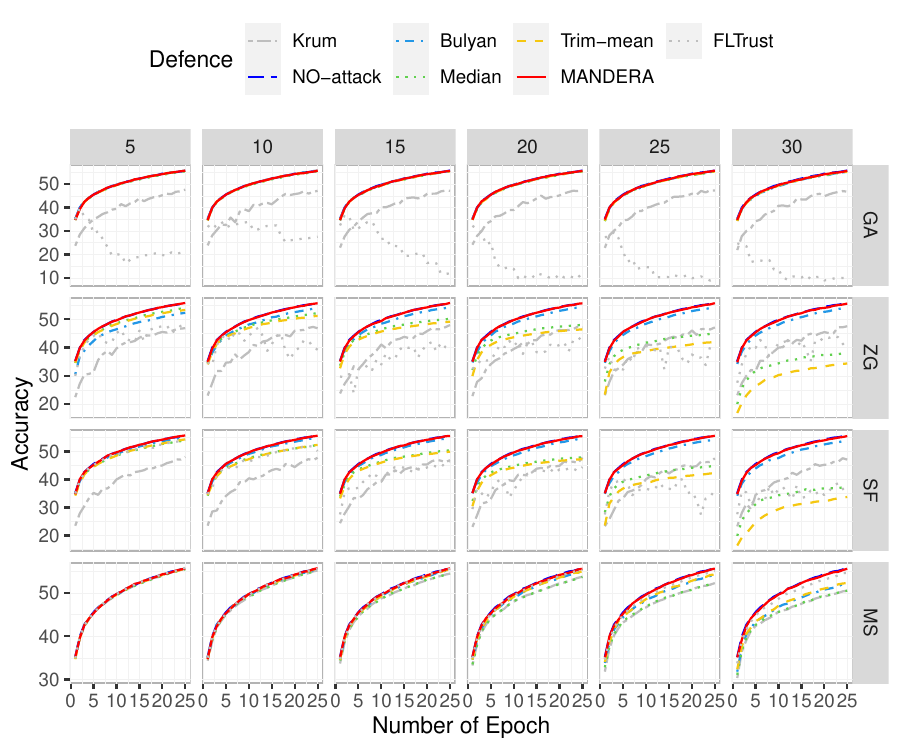}
    \includegraphics[width=11cm]{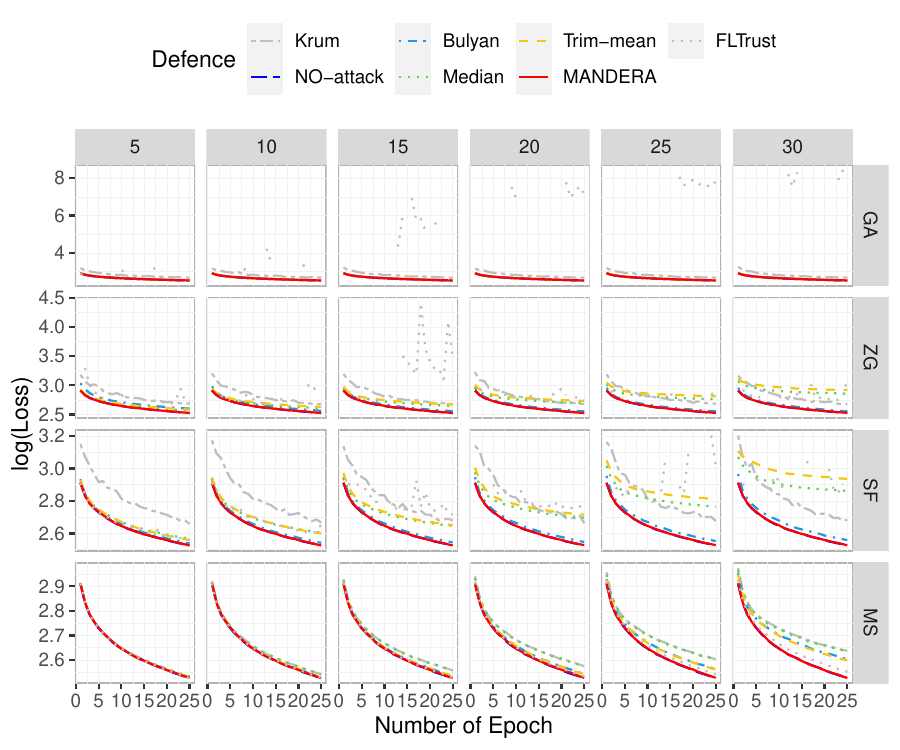}
    \caption{Trace plots of image classification accuracy and model log-loss of different defense methods on the CIFAR-10 dataset.}
    \label{fig:CIFAR10_TracePlot}
\end{figure}

Figure~\ref{fig:CIFAR_DetectionAccuracy} shows the metrics of MANDERA on the CIFAR-10 dataset. 
Table~\ref{tab:CIFAR_Accuracy} shows the model accuracy at 25th epoch with $n_m$ denoting the number of malicious nodes among 100 nodes for CIFAR-10 dataset.
 Figure~\ref{fig:CIFAR10_TracePlot} shows the trace plot of accuracy and loss of different defense methods on the CIFAR-10 dataset.

\begin{table}[tb]
\centering
\caption{CIFAR-10 model accuracy at 25th epoch with $n_m$ denoting the number of malicious nodes among 100 nodes. The {\bf bold} highlights the best defense strategy under attack. The baseline accuracy when no attack is conducted is \textbf{55.78}.}
\label{tab:CIFAR_Accuracy}
\resizebox{0.8\linewidth}{!}{  
\begin{tabular}{clcccccc}
  \toprule
Attack & Defense & $n_m=5$ & $n_m=10$ & $n_m=15$ & $n_m=20$ & $n_m=25$ & $n_m=30$ \\ \midrule
\multirow{6}{*}{GA}   
   & Median & 55.47 & 55.53 & 55.47 & 55.40 & 55.29 & 55.22 \\ 
   & Trim-mean & \textbf{55.77} & \textbf{55.72} & 55.56 & 55.50 & 55.43 & 55.31 \\ 
   & FLTrust & 19.66 & 27.54 & 11.99 & 9.21 & 9.73 & $\ $9.96 \\ \cline{2-8}
    & Krum & 47.66 & 47.16 & 47.18 & 47.26 & 47.25 & 46.77 \\ 
   & Bulyan & 55.69 & 55.85 & \textbf{55.67} & 55.63 & 55.46 & 55.22 \\
   & MANDERA & 55.74 & 55.69 & 55.63 & \textbf{55.65} & \textbf{55.76} & \textbf{55.69} \\
   \midrule
   
  \multirow{6}{*}{ZG}  
   & Median & 54.06 & 52.18 & 50.18 & 48.01 & 44.89 & 38.08 \\ 
   & Trim-mean & 53.34 & 51.22 & 49.14 & 46.45 & 42.02 & 34.36 \\ 
   & FLTrust & 48.05 & 39.21 & 39.44 & 44.25 & 40.27 & 39.49 \\ \cline{2-8}
   & Krum & 46.85 & 46.84 & 47.96 & 47.13 & 47.12 & 47.53 \\ 
   & Bulyan & 52.30 & 53.87 & 54.28 & 54.36 & 54.35 & 54.10 \\ 
   & MANDERA & \textbf{55.77} & \textbf{55.69} & \textbf{55.78} & \textbf{55.65 }& \textbf{55.72} & \textbf{55.56} \\ 
   \midrule
   
  \multirow{6}{*}{SF}  
   & Median & 53.96 & 52.29 & 50.49 & 47.89 & 44.93 & 37.22 \\ 
   & Trim-mean & 54.37 & 52.40 & 49.97 & 47.30 & 42.32 & 33.76 \\ 
   & FLTrust & 54.18 & 50.21 & 46.39 & 44.45 & 36.19 & 34.39 \\ \cline{2-8}
   & Krum & 48.11 & 47.79 & 46.93 & 47.89 & 47.59 & 47.13 \\ 
   & Bulyan & 55.30 & 54.99 & 54.86 & 54.68 & 54.43 & 54.05 \\ 
   & MANDERA & \textbf{55.78} & \textbf{55.69} & \textbf{55.62} & \textbf{55.55} &\textbf{ 55.67 }& \textbf{55.56} \\ 
   \midrule
   
  \multirow{6}{*}{MS}  
   & Median & 55.47 & 55.20 & 54.55 & 53.72 & 52.17 & 50.55 \\ 
   & Trim-mean & 55.64 & 55.59 & 55.38 & 55.09 & 54.29 & 52.32 \\ 
   & FLTrust & \textbf{55.81} & 55.64 & 55.62 & 55.42 & 55.09 & 54.65 \\ \cline{2-8}
   & Krum & 55.60 & 55.23 & 54.51 & 53.79 & 52.31 & 50.54 \\ 
   & Bulyan & 55.68 & 55.62 & 55.37 & 54.98 & 54.26 & 52.10 \\ 
   & MANDERA & 55.65 & \textbf{55.77} & \textbf{55.72} & \textbf{55.62} & \textbf{55.66} & \textbf{55.63} \\
   \bottomrule
\end{tabular}
}
\end{table}

\newpage
\subsection{MANDERA performance with alternative clustering algorithms}
\label{app:clustering}

\begin{figure*}[hb]
    \centering
    \begin{subfigure}[b]{0.54\linewidth}
    \includegraphics[width=\linewidth]{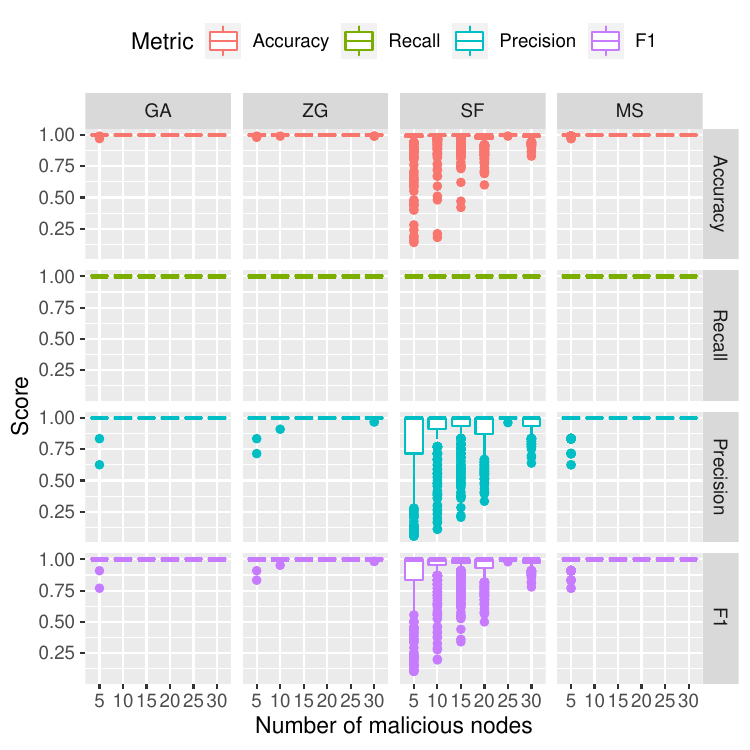}
    \caption{Clustering based on the Gaussian mixture model.}
    \label{fig:fashion.gmm.metric}
    \end{subfigure}
    \begin{subfigure}[b]{0.54\linewidth}
    \includegraphics[width=\linewidth]{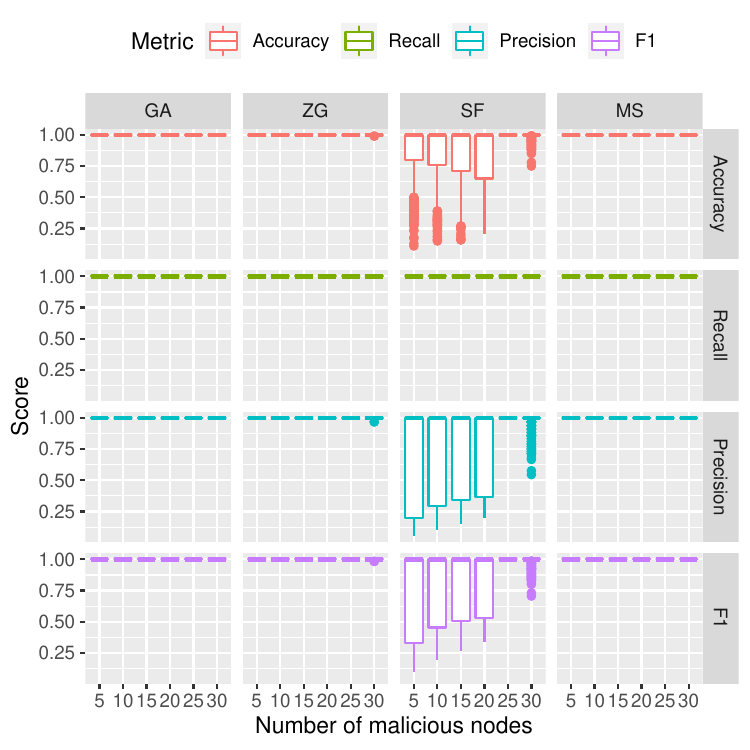}
    \caption{Hierarchical clustering.}
    \label{fig:fashion.hc.metric}
    \end{subfigure} 
    \caption{Classification performance of our proposed approach MANDERA (Algorithm 1) with other  clustering algorithms under four types of attack for FASHION-MNIST data. 
    }
    \label{fig:fashion.gmm.and.hc.metric}
\end{figure*}

In this section, Figure~\ref{fig:fashion.gmm.and.hc.metric} demonstrate that the discriminating performance of MANDERA when hierarchical clustering and Gaussian mixture models are used in-place of K-means for FASHION-MNIST data set remain robust.

\subsection{Impact of hyperparameters}

\begin{figure}[h]
    \centering
    \includegraphics[width=0.7\linewidth]{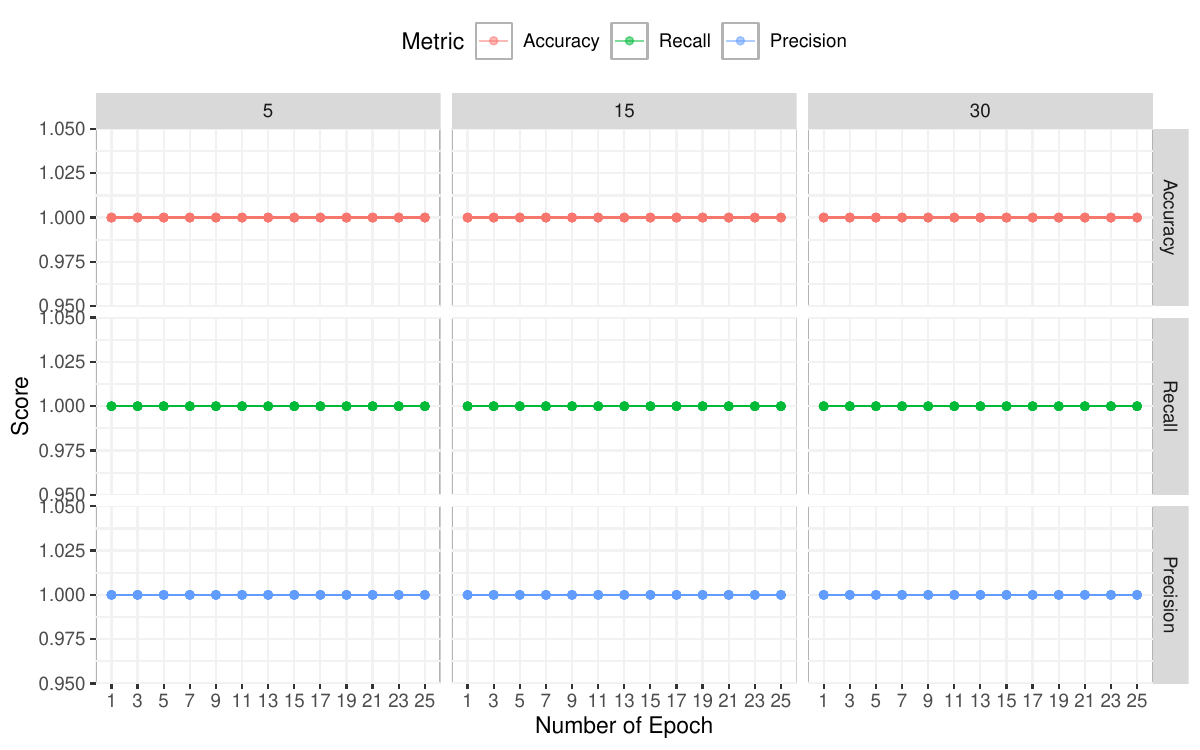}
    \caption{Performance of MANDERA on malicious node detection in the MNIST dataset under Gaussian attack.}
    \label{fig:MNIST_mandera}
\end{figure}

To investigate the impact of hyperparameter specification, we have conducted a few experiments where hyperparameters (e.g., learning rate and momentum) were specified at different values.
Firstly, we demonstrate in Figure \ref{fig:MNIST_mandera} how the performance of malicious node detection by MANDERA (i.e., accuracy, recall, precision) evolves along the model training process for the MNIST dataset under a Gaussian attack. Figure \ref{fig:MNIST_mandera} is obtained under our default hyperparameters. We can clearly see from the figure that the performance of MANDERA on malicious node detection stays unchanged when the number of epoch increases from 1 (unconverged stage) to 25 (converged stage). It shows that the status of convergence of the global model does not impact the proposed MANDERA. When we change the hyperameters, including the learning rate and the momentum, the performance of MANDERA is the same. We just omit the figures here.\\
Secondly, we show the impact of hyperparameters on the global model performance on image classification.
The results are demonstrated in Figure \ref{fig:MNIST_hyperpara} 
below, from which we can see that different hyperparameter specifications resulted in very similar performance of MANDERA.
In other words, MANDERA is not sensitive at all to the specification of hyperparameters.
This is because although hyperparameters have big impacts on the training process of a deep learning model, especially speed of convergence, they have little influence on MANDERA since MANDERA does not rely on the convergence of the global model to identify malicious nodes. 
This phenomenon suggests that MANDERA is universally effective in detecting malicious nodes at all stages of the model training process. 
Such a property of MANDERA roots in its working mechanism as described in Section 3 of the main texts.
In the revised manuscript, we have added a few sentences in Section 4 to discuss the impact of hyperparameter specification on MANDERA.

\begin{figure}[H]
    \centering
    \begin{subfigure} {0.48\linewidth}
    \centering
        \includegraphics[width=\linewidth]{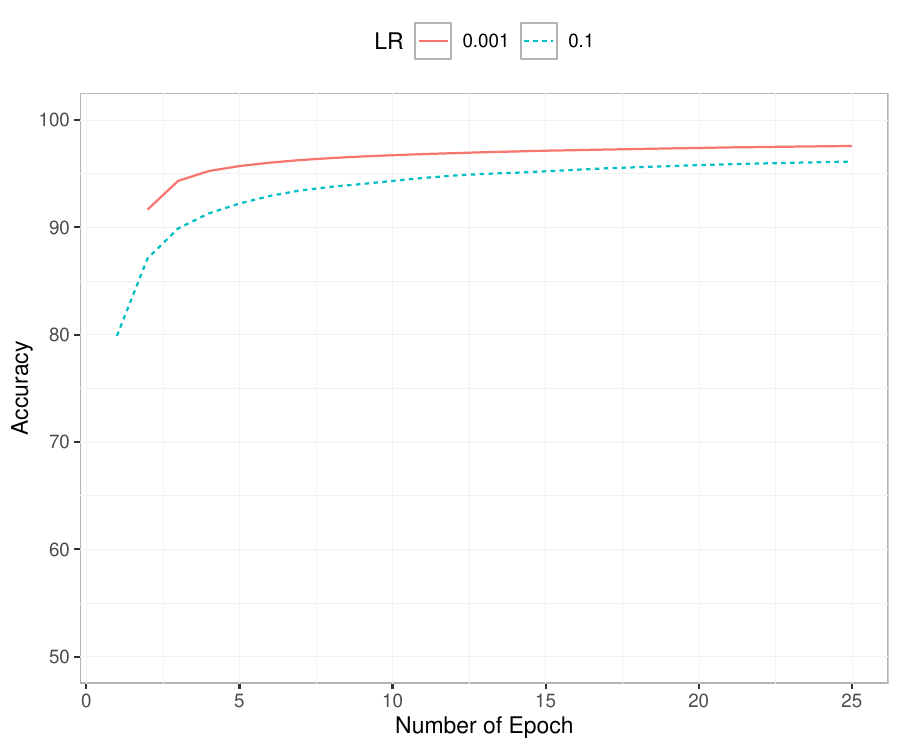}
        \caption{}
    \end{subfigure}
\hfill
\begin{subfigure} {0.48\linewidth}
\centering
    \includegraphics[width=\linewidth]{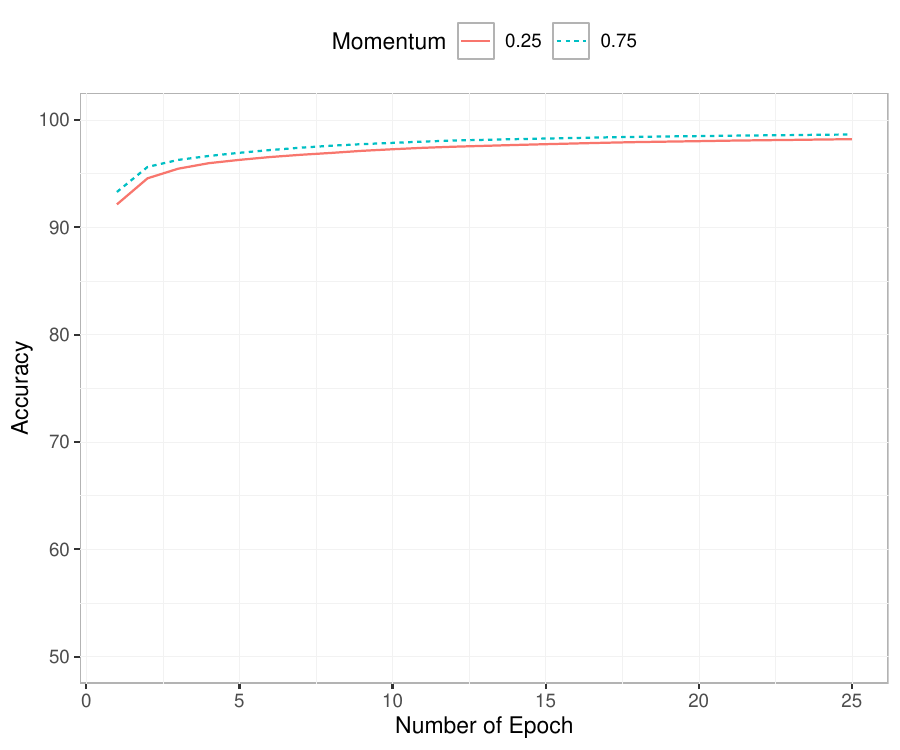}
    \caption{}
\end{subfigure}
    \caption{Image classification accuracy of the global model corrected by MANDERA under different specifications of hyperparameters for the MNIST dataset with Gaussian attack. (a) Comparison of two specifications of learning rate (LR). (b) Comparison of two specifications of momentum. All the other hyperparameters kept unchanged in the comparisons.}
    \label{fig:MNIST_hyperpara}
\end{figure}
\bibliography{ranking}
\bibliographystyle{imsart-nameyear}